\documentclass[11pt]{article}

\usepackage[final]{acl}

\usepackage{times}
\usepackage{latexsym}

\usepackage[T1]{fontenc}

\usepackage[utf8]{inputenc}

\usepackage{microtype}

\usepackage{inconsolata}
\usepackage{graphicx}
\usepackage{makecell}
\usepackage{tabularx}
\usepackage{array}
\usepackage{booktabs}
\usepackage{caption}
\usepackage{multirow}
\usepackage{stfloats}
\usepackage{amsmath}
\usepackage{pifont}
\usepackage{xcolor}
\usepackage{soul}

\definecolor{mygreen}{HTML}{28A745} 
\definecolor{myred}{HTML}{DC3545}   

\newcommand{\cmark}{\color{mygreen}\ding{51}} 
\newcommand{\xmark}{\color{myred}\ding{55}} 

\usepackage{amsfonts}
\usepackage[most]{tcolorbox}
\usepackage{enumitem}
\usepackage{float}
\usepackage{colortbl}
\usepackage{diagbox}

\usepackage{xspace}
\usepackage{placeins}
\usepackage{capt-of}
\usepackage{etoolbox}
\usepackage{cuted}

\newtcolorbox{PromptBox}[2][]{
    enhanced,
    breakable,             
    colback=gray!5,        
    colframe=black!70,     
    boxrule=0.8pt,
    arc=3pt,
    fonttitle=\bfseries\large,
    title={#2},            
    fontupper=\small\ttfamily, 
    #1
}

\newcommand{\dname}{\textsc{DeepChart}\xspace}

\title{\dname: How Far are LLMs from Faithful \\ Data-Science Chart Generation?}

\author{
  \textbf{Jiahui Tang\textsuperscript{1}},
  \textbf{Kuicai Dong\textsuperscript{2}},
  \textbf{Dexun Li\textsuperscript{2}},
  \textbf{Hongchao Gu\textsuperscript{1}},
  \textbf{Haocheng Yu\textsuperscript{1}}
  \\
  \textbf{Wei Han\textsuperscript{2}},
  \textbf{Chen Zhang\textsuperscript{2}},
  \textbf{Yong Liu\textsuperscript{2}},
  \textbf{Hao Wang\textsuperscript{1}},
  \textbf{Enhong Chen\textsuperscript{1}}
  \\
  {\normalfont\textsuperscript{1}University of Science and Technology of China}
  \\
  {\normalfont\textsuperscript{2}Huawei Technologies Co., Ltd.}
}

\begin{document}

\maketitle



\begin{abstract}

Faithful chart generation in real-world data-science workflows requires grounding visualizations in scattered evidence, computing chart-ready quantities, and rendering them accurately.
Modern LLMs can produce visually plausible, instruction-compliant charts, yet data-level hallucinations remain difficult to detect in long, noisy, and multimodal contexts.
To measure this gap, we introduce \dname, an expert-annotated benchmark of 1,482 task-conditioned chart-generation instances drawn from real-world scientific papers, financial filings, and ecosystem reports.
\dname formulates chart generation as an Extract--Reason--Visualize pipeline and evaluates source-data extraction, derived-data reasoning, and chart rendering stage by stage.
Experiments with state-of-the-art models show that visually plausible charts often conceal data-level hallucinations, with extraction and reasoning errors common in realistic long and multimodal settings.
These findings suggest that larger context windows alone are insufficient; faithful chart generation also requires reliable evidence extraction and quantitative reasoning before rendering.
Our benchmark and associated resources are available at
\url{https://github.com/tangdouer1005/DeepChart}.

\end{abstract}

\begin{figure*}[t]
    \centering
    \includegraphics[width=1\textwidth]{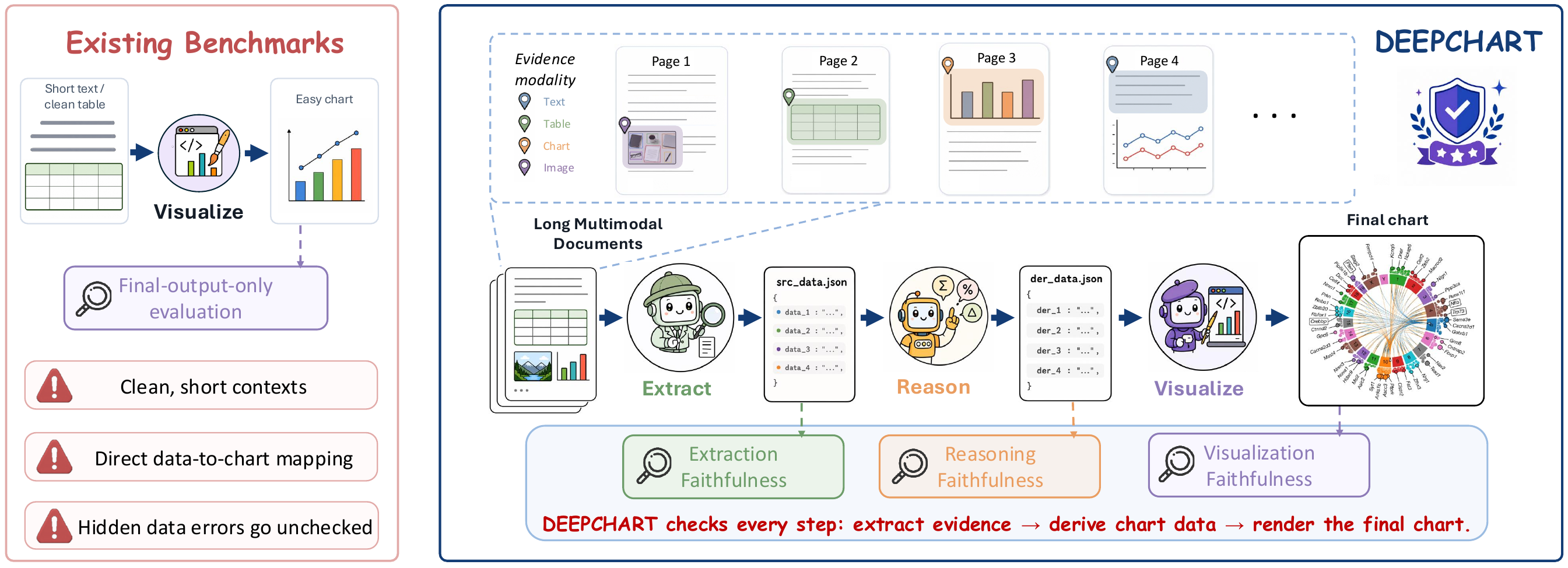}
    \caption{\textbf{\dname task overview.} Compared with visualize-only chart benchmarks, \dname evaluates chart generation as an Extract--Reason--Visualize pipeline over long multimodal documents.}
    \label{fig:teaser}
\end{figure*}

\section{Introduction}
Charts are a primary medium for communicating data-science findings~\cite{munzner2014visualization}.
In real-world workflows, however, producing a faithful chart is rarely a matter of plotting a ready-made table.
An analyst must sift through extensive, heterogeneous documents, extract scattered data points, apply non-trivial transformations, and only then render the result as a visualization.
As LLMs are increasingly used as data-analysis agents~\cite{singh2025agentic,jin2025search}, their outputs must move beyond textual answers to communicate derived findings through faithful visualizations~\cite{huang2025deepresearchagentssystematic, dong2025benchmarking}.
This raises a challenge that is easy to overlook: \emph{the data faithfulness of a chart cannot be observed from the chart itself}.
A figure rendered from mis-extracted or mis-computed numbers can still look visually plausible. 
The act of rendering conceals the errors that precede it.
We call such surviving errors \emph{hidden hallucinations}: subtle mistakes in extraction or reasoning that pass through rendering and evade conventional, appearance-based evaluation.

Current chart-generation benchmarks are ill-equipped to expose hidden hallucinations because they often map clean inputs directly to final charts and score only the endpoint.
As summarized in Table~\ref{tab:comparison_bench}, this gap appears in three aspects.
\textbf{(1)~\textit{Pre-digested inputs.}}
Existing benchmarks supply clean tables or short snippets rather than the large, heterogeneous, multimodal documents of real analysis, removing or simplifying the stages where hidden hallucinations originate.
\textbf{(2)~\textit{Missing intermediate data references.}}
Generation is treated as a one-shot prompt-to-chart mapping, with no reference source values or derived quantities for inspecting what happened before rendering.
\textbf{(3)~\textit{Endpoint evaluation.}}
Consequently, final-chart scoring can neither verify the underlying numbers nor localize failures to extraction, reasoning, or rendering.

To detect hidden hallucinations, we model chart generation as a \textbf{multi-stage analytical pipeline} that captures key data-science bottlenecks, as illustrated in Figure~\ref{fig:teaser}.
Inspired by the classical ETL (Extract$\to$Transform$\to$Load) workflow~\cite{khan2024overview}, we decompose this pipeline into \textbf{\textit{Extract}} (retrieve intent-relevant evidence), \textbf{\textit{Reason}} (derive chart-ready quantities), and \textbf{\textit{Visualize}} (render faithful charts).
This decomposition enables failure isolation across extraction, reasoning, and visualization.

Building on this formulation, we introduce \dname, an expert-annotated benchmark for faithful data-science chart generation.
\dname contains task-conditioned context-query instances drawn from real-world documents across text and multimodal protocols.
Its text inputs average 220.6K tokens, its multimodal report inputs average 218 pages, and the average instance draws on 317.6 context-sourced data points.
Beyond direct data-to-visual mapping, each instance requires the \textit{ERV} pipeline: models must extract intent-relevant evidence, derive chart-ready data, and then render the chart.
We instantiate this design across Academic-Normal/Long, Finance-Normal/Long/Ultra-Long, and Ecosystem report-level multimodal settings, covering both context-scale variation and native multimodal document inputs.
To expose hidden hallucinations, \dname provides stage-by-stage evaluation aligned with the \textit{ERV} pipeline: \textbf{\textit{Source Data Fidelity}} for extraction, \textbf{\textit{Derived Data Fidelity}} for reasoning, and \textbf{\textit{Visual Accuracy Score}~(VAS)} for rendered-chart faithfulness.

Zero-shot evaluation on \dname shows that (1) visually plausible charts often conceal data-level hallucinations, (2) models struggle with both Extract and Reason stages, and (3) full-document inputs further degrade end-to-end faithfulness, suggesting that larger context windows alone are insufficient.
In summary, our contributions are:
\begin{itemize}[leftmargin=*, itemsep=0.0em, topsep=0.0em]
    \item \textbf{Task formulation.}
    We formalize data-science chart generation as an \textit{ERV} pipeline, exposing the coupled challenges of evidence retrieval, analytical reasoning, and faithful rendering.

    \item \textbf{\dname benchmark.}
    We release an expert-annotated benchmark over long-text and report-level multimodal contexts, with references and metrics for source values, derived quantities, and final charts.

    \item \textbf{Empirical findings.}
    We show that SOTA models suffer from hidden hallucinations, persistent extraction and reasoning challenges, and degraded faithfulness under full-document inputs.
\end{itemize}

\begin{table*}[t]
    \centering
    \resizebox{\textwidth}{!}{
    \scriptsize
    \renewcommand{\arraystretch}{0.95}
    \setlength{\tabcolsep}{4pt}

    \begin{tabular}{l *{7}{c}}
        \toprule
        \multirow{2}{*}{\textbf{Method}}
        & \multicolumn{3}{c}{\textbf{Input Setting}}
        & \multicolumn{2}{c}{\textbf{Task Requirement}}
        & \multicolumn{2}{c}{\textbf{Evaluation Granularity}} \\

        \cmidrule(lr){2-4}
        \cmidrule(lr){5-6}
        \cmidrule(lr){7-8}

        & \textbf{Context}
        & \textbf{Modality}
        & \textbf{Real-doc}
        & \textbf{Retrieval}
        & \textbf{Reasoning}
        & \textbf{Visual}
        & \textbf{Stage-by-stage} \\
        \midrule

        Text2Chart31 \cite{pesaran-zadeh-etal-2024-text2chart31}
        & <16K & Txt+Tab & \xmark
        & \xmark & \xmark
        & \cmark & \xmark \\

        MatPlotBench \cite{MatPlotBench}
        & <32K & Tab & \xmark
        & \xmark & \cmark
        & \cmark & \xmark \\

        Text2Vis \cite{rahman-etal-2025-text2vis}
        & <2K & Tab & \xmark
        & \cmark & \cmark
        & \cmark & \xmark \\

        C$^2$/ChartUIE-8K \cite{koh2025c2scalableautofeedbackllmbased}
        & <32K & Tab & \xmark
        & \xmark & \cmark
        & \cmark & \xmark \\

        Doc2Chart \cite{jain-etal-2025-doc2chart}
        & <32K$^{*}$ & Txt+Tab & \cmark
        & \cmark & \cmark
        & \cmark & \xmark \\

        Infogen \cite{ghosh-etal-2025-infogen}
        & <2K$^{*}$ & Txt & \xmark
        & \cmark & \xmark
        & \cmark & \xmark \\

        PlotCraft \cite{zhang2026plotcraftpushinglimitsllms}
        & <128K$^{*}$ & Tab & \cmark
        & \xmark & \cmark
        & \cmark & \xmark \\

        \midrule

        \dname
        & \makecell{220K tokens / \\ 218 pages .avg}& Txt+Tab+Img & \cmark
        & \cmark & \cmark
        & \cmark & \cmark \\

        \bottomrule
    \end{tabular}
    }
    \caption{
\textbf{Comparison with existing chart-generation benchmarks.}
\textbf{Context} reports approximate input scale; starred values are estimated when complete statistics are unavailable.
\textbf{Retrieval} means locating task-relevant data in complex contexts; 
\textbf{Reasoning} means deriving chart-ready quantities from extracted data.
\textbf{Real-doc} indicates contexts are constructed from real-world documents, 
and \textbf{Stage-by-stage} indicates evaluation of intermediate data rather than only the final chart.
}
    \label{tab:comparison_bench}
\end{table*}

\section{Task Definition}

Given a long, heterogeneous analytical context $C$ and a natural-language chart intent $Q$, the task is to generate a faithful data-science chart. The context may contain textual descriptions, tables, figures, or document pages, and the chart intent specifies the analytical goal and desired visualization. Unlike direct prompt-to-chart generation, the pipeline is expected to expose the data path that supports the final visualization: the source values extracted from the context, the chart-ready quantities derived from those values, and the executable visualization that renders them.

We formulate this process as an Extract--Reason--Visualize (ERV) pipeline:
\begin{align}
D_{\mathrm{src}} &\leftarrow \mathrm{Extract}(C, Q), \\
D_{\mathrm{der}} &\leftarrow \mathrm{Reason}(D_{\mathrm{src}}, Q), \\
(P, G) &\leftarrow \mathrm{Visualize}(D_{\mathrm{src}}, D_{\mathrm{der}}, Q).
\end{align}
Here, $D_{\mathrm{src}}$ denotes extracted source data, $D_{\mathrm{der}}$ denotes derived chart-ready data, $P$ denotes the executable chart-generation program, and $G$ denotes the rendered chart.

We define the auditable intermediate state as $J=(D_{\mathrm{src}}, D_{\mathrm{der}})$. 
Exposing $J$ makes a generated chart traceable: we can verify whether it is grounded in the correct source values and whether those values are transformed into the correct chart-ready quantities before rendering.
This enables stage-by-stage evaluation across the Extract, Reason, and Visualize stages, rather than assessing only the final chart.
With this design, \dname can distinguish extraction, reasoning, and visualization failures, thereby exposing hidden hallucinations.

\section{Benchmark: \dname}

\label{sec:benchmark_construction}

To capture diverse real-world chart-generation workflows, \dname spans three domains with distinct sources, modalities, and bottlenecks. 
The \textbf{Academic} domain uses scientific papers and supplementary materials, stressing evidence localization, statistical reasoning, and faithful scientific visualization. 
The \textbf{Finance} domain uses long, table-heavy company 10-K filings, stressing cross-table aggregation and multi-hop derivation of financial metrics. 
The \textbf{Ecosystem} domain uses market intelligence and startup ecosystem reports that mix textual, tabular, and visual evidence, stressing multimodal evidence integration and reasoning. 
These domains are selected to vary both input scale and evidence form, from text-centered scientific and financial documents to page-level multimodal reports.
Together, these domains cover three recurring bottlenecks: statistical derivation, long-context aggregation, and multimodal evidence integration.

\begin{figure*}[t]
    \centering
    \includegraphics[width=1\textwidth]{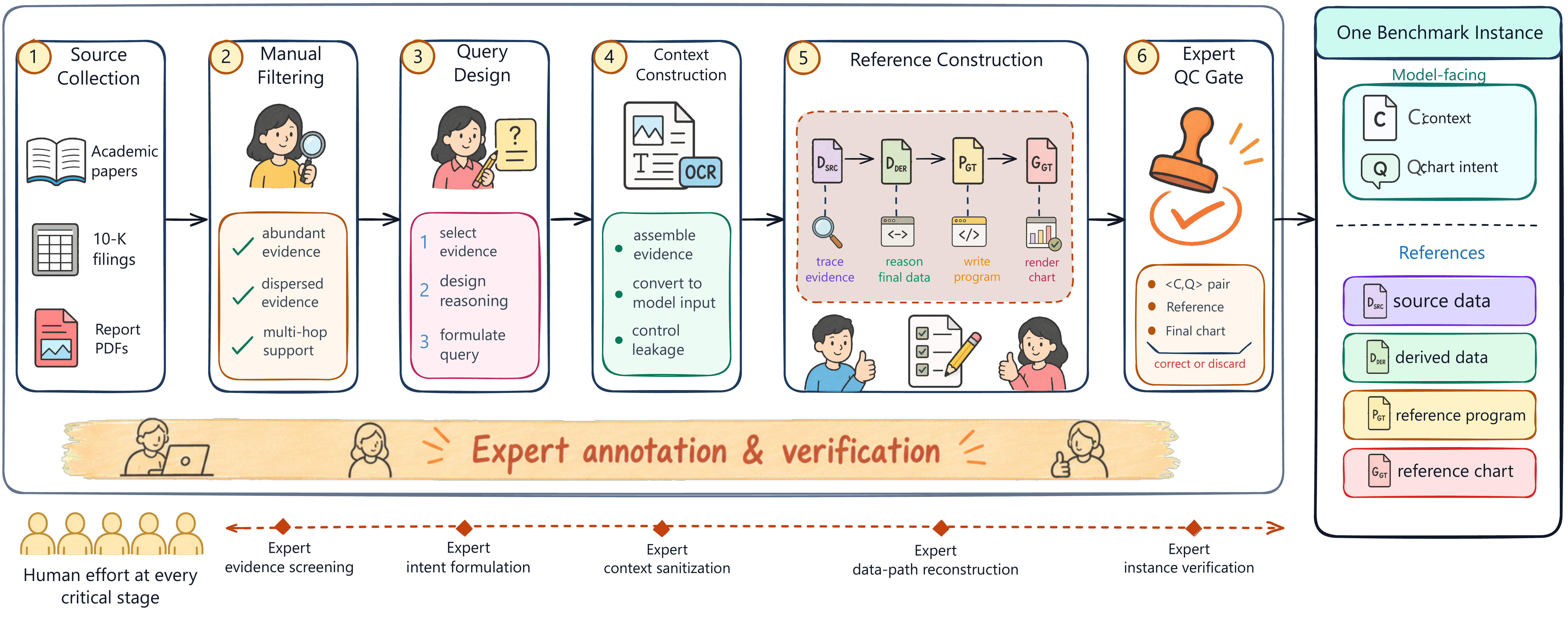}
    \caption{\textbf{Expert-verified benchmark construction.} Experts validate source selection, query design, reference data, executable code, and rendered charts. Each final instance contains a model-facing context-query pair and hidden references for stage-wise evaluation of extraction, reasoning, and visualization.}
    \label{fig:pipeline}
\end{figure*}

Each \dname instance consists of model-facing inputs $\langle C,Q\rangle$ and hidden evaluation references: an auditable intermediate state $J_{\mathrm{GT}}=(D_{\mathrm{src}},D_{\mathrm{der}})$, an executable reference program $P_{\mathrm{GT}}$, and a rendered reference chart $G_{\mathrm{GT}}$. This design entails two construction requirements: inputs must preserve the complexity of real-world data-science workflows, where evidence is scattered across long, heterogeneous, and sometimes multimodal documents; references must expose an auditable data path from source evidence to derived quantities and the final chart. Figure~\ref{fig:pipeline} summarizes our expert-verified pipeline, from source collection and context-query construction to reference construction and quality control.







\subsection{Task Input Construction}

We construct the task input $\langle C,Q\rangle$ in three steps: collecting and filtering source documents, converting them into standardized contexts $C$, and pairing each context with a chart intent $Q$.

\textbf{Source collection and filtering.}
We manually collect source documents from publicly available repositories and official filings. For the Academic domain, we retrieved academic papers along with their supplementary materials. Finance domain documents consist of company 10-K filings in PDF format. The Ecosystem domain includes market intelligence and startup ecosystem reports. 

We manually filter these sources to remove low-quality and unsuitable documents. 
Specifically, we exclude documents that are too short, lack sufficient quantitative information, or contain evidence that is insufficiently dispersed. 
The remaining sources thus provide the complexity required to construct challenging \dname tasks.

\textbf{Query construction.}
From each retained source, experts construct standardized chart intents Q corresponding to the evidence provided in the document. 
In the Academic domain, experts first identify the target charts in the papers. 
They then formulate queries based on the chart and its surrounding context. 
For the Finance and Ecosystem domains, experts systematically identify all available quantitative evidence in each report, including direct indicators, tables, textual statistics, and visual elements. 
Based on these evidence pools, they formulate analytical queries that require deriving quantities through multi-hop reasoning prior to visualization. 
During query design, experts ensure that each task draws on evidence that is abundant, widely distributed, and, when applicable, spans multiple modalities, tables, or documents.

\textbf{Context construction.}
Given a query $Q$, we build a task-conditioned context $C$ that is answerable, leakage-controlled, and directly consumable by LLMs. Specifically, $C$ must contain the evidence needed to solve $Q$, exclude target artifacts such as the reference chart, and be converted from the original PDFs into text or images. 

For text-domain tasks, documents are converted to text using PaddleOCR-VL~\cite{cui2025paddleocrvlboostingmultilingualdocument}, and we further create context-scale variants to probe how context length affects the \textit{ERV}  pipeline.

In Academic, the original papers contain target charts. After OCR conversion, the paper is text-only, so the chart image is not exposed and does not cause data leakage. To keep the task answerable, experts then identify the experimental data corresponding to each target chart and insert them as a structured data block at the original chart location. The \textit{Normal} setting contains the experimental data required for chart construction, while the \textit{Long} setting adds the full paper text.

In Finance, experts first select the company filings required by each query. Since a 10-K filing often reports indicators for two or three fiscal years, we remove redundant filings with overlapping coverage while ensuring that the query remains answerable. The \textit{Normal} setting keeps only the tables that contain the required source values. The \textit{Long} setting adds text from the 20 rows above and below each table. The \textit{Ultra-Long} setting uses all selected full 10-K filings.

For multimodal-domain Ecosystem, PDF pages are converted into images, forming a native report-level multimodal context for visual models.

The resulting benchmark contains 1,482 context-query instances. Text-domain queries may be paired with multiple context-scale variants, while the query and evaluation references remain fixed. Detailed statistics are reported in Appendix~\ref{app:benchmark_stats}.

\subsection{Auditable Ground Truth Construction}

\dname evaluation requires references beyond the final chart. 
For each task $\langle C,Q\rangle$, we construct references for every stage of the Extract--Reason--Visualize pipeline. The source data $D_{\mathrm{src}}$ captures the output of the \textbf{Extract} step, and the derived data $D_{\mathrm{der}}$ captures the output of the \textbf{Reason} step. The executable program $P_{\mathrm{GT}}$ and rendered chart $G_{\mathrm{GT}}$ correspond to the \textbf{Visualize} step.

\textbf{Evidence Extraction.}
Given the context $C$ and query $Q$, expert annotators identify the source values required to answer the task and organize them as $D_{\mathrm{src}}$. These values may originate from text, tables, or visual evidence, as appropriate to the domain.

\textbf{Data Reasoning.}  
Based on the requirements specified by $Q$, expert annotators implement a Python script to transform $D_{\mathrm{src}}$ into derived quantities $D_{\mathrm{der}}$. This script encodes the necessary reasoning (e.g., aggregation, normalization, ratio calculations, delta computation, grouping, and cross-year comparisons), making the derivation from source to derived values fully reproducible.

\textbf{Reference Chart Construction.}  
Using $D_{\mathrm{src}}$, $D_{\mathrm{der}}$, expert annotators create an executable reference program $P_{\mathrm{GT}}$. Executing $P_{\mathrm{GT}}$ produces the reference chart $G_{\mathrm{GT}}$, which is verified to ensure that it can solve the task $\langle C,Q\rangle$ faithfully.

LLMs may assist in drafting code, but only as a drafting tool. They are not treated as authoritative sources for ground truth. Expert annotators verify source data, derived data, code, and rendered charts before inclusion in the benchmark.

\subsection{Quality Control}

\dname was constructed and verified by five expert annotators whose expertise spans data visualization, document understanding, quantitative analysis, scientific literature, and financial-report analysis. 
Here, \textit{expert annotator} denotes a member of this team; \textit{primary annotator}, \textit{independent verifier}, and \textit{senior reviewer} denote QC roles.

Each instance is authored by a primary expert annotator and independently reviewed by an expert verifier across input, data, code, and chart. 
The verifier checks that $\langle C,Q\rangle$ is answerable and leakage-free, that $D_{\mathrm{src}}$ is evidence-supported, that $D_{\mathrm{der}}$ is reproducibly derived from $D_{\mathrm{src}}$, and that executing $P_{\mathrm{GT}}$ yields a chart $G_{\mathrm{GT}}$ consistent with the intended data and visual specification. 
Ambiguous cases are discussed with a senior expert reviewer; unverifiable instances are discarded, and correctable local issues are revised and rechecked.



\section{Evaluation Framework}
\label{sec:evaluation}

We evaluate faithful data-science chart generation with stage-by-stage metrics aligned with the \textit{ERV} pipeline. The framework reports Source Data Fidelity for extraction ($F_{1,\mathrm{src}}$), Derived Data Fidelity for reasoning ($F_{1,\mathrm{der}}$), Visual Accuracy Score (VAS) for visualization, and Execution Rate (ER) as an auxiliary measure of rendering success.




\textbf{Extract: Source Data Fidelity.}
$F_{1,\mathrm{src}}$ measures recovery of query-relevant numerical evidence by normalized numeric matching against reference source values, independent of JSON formatting.

\textbf{Reason: Derived Data Fidelity.}
$F_{1,\mathrm{der}}$ uses the same matching to evaluate chart-ready quantities derived from the source data, capturing errors in aggregation, normalization, ratio calculation, and cross-year comparison.

Both metrics operate at the value level: they extract normalized numeric leaves from predicted and reference JSON objects and perform one-to-one matching without key- or structure-level alignment. 
Thus, they measure numerical value recovery rather than full semantic alignment.

\textbf{Visualize: Visual Accuracy Score.}
For each benchmark instance, we generate and cache binary verification questions from the reference chart and query. The questions cover three aspects: instruction compliance, data-mapping topology, and presentation quality. A VLM judge answers these questions for each model-generated chart, and VAS is the average pass rate:
\begin{equation}
    \mathrm{VAS} =
    \frac{1}{N} \sum_{i=1}^{N}
    \mathbf{1}(\mathrm{Rubric}_i = \text{Pass}).
\end{equation}

\paragraph{Execution Rate (ER).}
ER is the percentage of test instances for which the generated visualization program successfully produces a valid chart image. 

Appendix~\ref{app:evaluation_details} provides implementation details for data-fidelity matching, VAS rubric generation, judge models, and execution checks; Appendix~\ref{app:vas_validation} reports VAS reliability validation.

\section{Experiments}

\subsection{Experimental Setup}

We evaluate \dname under two protocols. 
The text-domain protocol covers Academic and Finance, using the context-scale variants defined in Section~\ref{sec:benchmark_construction}: Academic-Normal/Long and Finance-Normal/Long for the main benchmark, with Finance-Ultra-Long evaluated separately as a stress test in Section~\ref{sec:ultra_long_context}. The multimodal-domain protocol covers Ecosystem, where models receive report pages as images and are evaluated under the native report-level setting.

For the text-domain protocol, we evaluate four proprietary models (GPT-5.2~\cite{gpt5.2}, Claude 4.5 Opus~\cite{ClaudeOpus4.5}, Gemini 3 Pro~\cite{gemini3pro}, and Gemini 3 Flash~\cite{gemini3flash}) and four open-weight models (DeepSeek-V3.2~\cite{deepseekv3.2}, GLM-4.7~\cite{glm4.7}, Qwen3-235B-A22B-thinking~\cite{qwen3}, and Kimi-K2-Thinking~\cite{kimik2thinking}). 
For the multimodal-domain protocol, we retain the same proprietary models and add four multimodal-capable models (Qwen3.6-Plus~\cite{qwen3}, Qwen3.6-Flash~\cite{qwen3}, Kimi-2.6~\cite{kimi25}, and Qwen3-VL-30B-A3B-Thinking~\cite{qwen3vl}).

All experiments are zero-shot. We report Source Data Fidelity ($F_{1,\mathrm{src}}$), Derived Data Fidelity ($F_{1,\mathrm{der}}$), Visual Accuracy Score (VAS), and Execution Rate (ER), as defined in Section~\ref{sec:evaluation}. Implementation details, prompt templates, model versions, and VAS validation are provided in Appendix~\ref{app:generation_details}.

\definecolor{graybg}{gray}{0.95}

\begin{table*}[t]
    \renewcommand\arraystretch{1.15} 
    \small
    \centering
    
    \resizebox{0.9\linewidth}{!}{%
    \begin{tabular}{l|cccc|cccc}
        \toprule
        \multirow{2}{*}{\diagbox[width=12em]{\textbf{Model}}{\textbf{Metric}}} 
        & \multicolumn{4}{c|}{\textbf{Finance}} 
        & \multicolumn{4}{c}{\textbf{Academic}} \\ 
        \cmidrule(lr){2-5}
        \cmidrule(lr){6-9}

         & \textbf{ER} & \textbf{$F_{1,\mathrm{src}}$} & \textbf{$F_{1,\mathrm{der}}$} & \textbf{VAS} 
         & \textbf{ER} & \textbf{$F_{1,\mathrm{src}}$} & \textbf{$F_{1,\mathrm{der}}$} & \textbf{VAS} \\ 
        \midrule
        
        \multicolumn{9}{c}{\cellcolor{graybg}\textbf{Normal Context Settings}} \\ 
        \midrule

        \multicolumn{9}{l}{\textit{\textbf{Proprietary Models}}} \\
        GPT-5.2             & 0.8415 & 0.4084 & 0.2948 & 0.5265 & 0.6685 & 0.7181 & 0.1994 & 0.4948 \\
        Claude 4.5 Opus   & \textbf{0.8908} & 0.4006 & 0.1771 & \textbf{0.5710} & 0.7697 & \textbf{0.7855} & 0.2273 & \textbf{0.6002} \\
        Gemini 3 Pro      & 0.6268 & \textbf{0.4984} & 0.4162 & 0.3982 & 0.3989 & 0.7544 & 0.2674 & 0.2992 \\
        Gemini 3 Flash    & 0.8556 & 0.4849 & \textbf{0.4769} & 0.5398 & 0.5730 & 0.7686 & 0.2681 & 0.4167 \\
        
        \multicolumn{9}{l}{\textit{\textbf{Open-weight Models}}} \\
        DeepSeek-V3.2       & 0.4225 & 0.2107 & 0.0825 & 0.2313 & 0.4775 & 0.6616 & 0.2373 & 0.3274 \\
        GLM-4.7           & 0.3415 & 0.3742 & 0.2453 & 0.1863 & 0.3315 & 0.6252 & 0.1800 & 0.2225 \\
        Kimi-K2           & 0.7606 & 0.4314 & 0.1930 & 0.4472 & 0.5562 & 0.5022 & 0.2179 & 0.3859 \\
        Qwen3-235B        & 0.8627 & 0.2738 & 0.1046 & 0.4686 & \textbf{0.8034} & 0.7123 & \textbf{0.2895} & 0.5291 \\

        \midrule
        
        \multicolumn{9}{c}{\cellcolor{graybg}\textbf{Long Context Settings}} \\ 
        \midrule

        \multicolumn{9}{l}{\textit{\textbf{Proprietary Models}}} \\
        GPT-5.2             & 0.8627 & 0.4158 & 0.3237 & 0.5387 & 0.6854 & 0.6066 & 0.1775 & 0.5238 \\
        Claude 4.5 Opus   & 0.8592 & 0.4127 & 0.2325 & 0.5330 & \textbf{0.7809} & 0.6163 & 0.1868 & \textbf{0.5945} \\
        Gemini 3 Pro      & 0.7606 & 0.4587 & 0.2915 & 0.4716 & 0.3820 & \textbf{0.6870} & 0.2158 & 0.2888 \\
        Gemini 3 Flash    & 0.8592 & \textbf{0.4817} & \textbf{0.4840} & \textbf{0.5474} & 0.5281 & 0.6422 & 0.2278 & 0.3938 \\

        \multicolumn{9}{l}{\textit{\textbf{Open-weight Models}}} \\
        DeepSeek-V3.2       & 0.5000 & 0.1936 & 0.0640 & 0.1557 & 0.3539 & 0.4401 & 0.2366 & 0.2450 \\
        GLM-4.7           & 0.3838 & 0.2486 & 0.1246 & 0.1739 & 0.4663 & 0.5521 & 0.2037 & 0.3170 \\
        Kimi-K2           & 0.5106 & 0.4268 & 0.1587 & 0.2993 & 0.6517 & 0.6617 & 0.1787 & 0.4696 \\
        Qwen3-235B        & \textbf{0.9085} & 0.2882 & 0.0560 & 0.3535 & 0.6854 & 0.5985 & \textbf{0.2402} & 0.4610 \\



        

        \bottomrule
    \end{tabular}
    }
\caption{\textbf{Text-domain results} on Academic and Finance under Normal and Long context settings.}

\label{tab:text_setting}
\end{table*}

\begin{table}[t]
    \renewcommand\arraystretch{1.15}
    \small
    \centering
    \resizebox{\columnwidth}{!}{%
    \begin{tabular}{l|cccc}
        \toprule
        \multicolumn{5}{c}{\cellcolor{graybg}\textbf{Ecosystem}} \\
        \textbf{Model} & \textbf{ER} & \textbf{$F_{1,\mathrm{src}}$} & \textbf{$F_{1,\mathrm{der}}$} & \textbf{VAS} \\
        \midrule

        \multicolumn{5}{l}{\textit{\textbf{Shared Models}}} \\
        GPT-5.2          & 0.7031 & 0.0207 & 0.0386 & 0.1589 \\
        Claude 4.5 Opus  & 0.9609 & 0.1792 & 0.2832 & 0.6963 \\
        Gemini 3 Pro     & \textbf{0.9844} & 0.1965 & \textbf{0.4529} & \textbf{0.7427} \\
        Gemini 3 Flash   & 0.6133 & \textbf{0.2167} & 0.4107 & 0.4055 \\

        \midrule
        \multicolumn{5}{l}{\textit{\textbf{Additional Models}}} \\
        Qwen3.6-Plus     & 0.9688 & 0.2122 & 0.4088 & 0.5984 \\
        Qwen3.6-Flash    & 0.8047 & 0.1472 & 0.2633 & 0.2387 \\
        Kimi-2.6         & 0.6875 & 0.1564 & 0.2796 & 0.4998 \\
        Qwen3-VL-30B & 0.5352 & 0.0648 & 0.0746 & 0.2358 \\

        \bottomrule
    \end{tabular}
    }
    \caption{\textbf{Multimodal-domain results} on Ecosystem under the native report-level setting.}
    \label{tab:multimodel_setting}
\end{table}

\subsection{Main Benchmark Results}

Tables~\ref{tab:text_setting} and~\ref{tab:multimodel_setting} report the main results under the text-domain and multimodal-domain protocols. Across both protocols, current models remain far from reliable faithful chart generation under our zero-shot ERV setting. The results breakdown reveals three patterns that output-only evaluation would obscure.
\textbf{(1) Executability does not guarantee visual faithfulness.}
In the Ecosystem domain, models achieve average ER (0.782), but average VAS is only 0.447. This shows that a generated program may execute successfully while the resulting chart still fails to satisfy the intended visual specification.
\textbf{(2) Visual faithfulness does not guarantee data faithfulness.}
Even when charts appear visually plausible, their underlying data path may be incorrect. In Ecosystem, average VAS (0.447) is substantially higher than average $F_{1,\mathrm{src}}$ (0.149) and average $F_{1,\mathrm{der}}$ (0.276), showing that visually acceptable charts can still rely on incorrect source or derived data.
\textbf{(3) Reliable extraction does not guarantee correct reasoning.}
In Academic, models recover source values relatively well, with average $F_{1,\mathrm{src}}$ (0.691/0.601 under Normal/Long), but average $F_{1,\mathrm{der}}$ remains much lower (0.236/0.208). This indicates that extracting relevant evidence does not ensure correct chart-ready derivation.
Together, these results show why stage-by-stage evaluation is necessary. By separately measuring execution, visualization, source extraction, and derived reasoning, \dname exposes hidden hallucinations that final-output-only evaluations would miss.

The results also reveal domain-specific bottlenecks. 
\textbf{Finance} stresses long-context aggregation: moving from Normal to Long produces modest but consistent drops in average VAS (0.421 to 0.384) and average $F_{1,\mathrm{der}}$ (0.249 to 0.217), suggesting that additional filing context increases the difficulty of retrieval and derivation. 
\textbf{Academic} stresses scientific derivation: models recover source values relatively well, with $F_{1,\mathrm{src}}$ reaching up to 0.785, but $F_{1,\mathrm{der}}$ remains below 0.240, showing difficulty in transforming raw experimental evidence into chart-ready quantities. 
\textbf{Ecosystem} stresses multimodal grounding and reasoning: despite average ER (0.782), models achieve only average $F_{1,\mathrm{src}}$ (0.149) and average $F_{1,\mathrm{der}}$ (0.276). 
Together, these patterns show that \dname evaluates long-context retrieval, scientific derivation, multimodal grounding, and rendering within a unified chart-generation benchmark.


\subsection{Impact of Ultra-Long Context}
\label{sec:ultra_long_context}

\begin{table}[t]
    \renewcommand\arraystretch{1.15}
    \small
    \centering
    \resizebox{\columnwidth}{!}{%
    \begin{tabular}{l|cccc}
        \toprule
        \multicolumn{5}{c}{\cellcolor{graybg}\textbf{Finance-Ultra-Long}} \\
        \midrule
        \textbf{Model} & \textbf{ER} & \textbf{$F_{1,\mathrm{src}}$} & \textbf{$F_{1,\mathrm{der}}$} & \textbf{VAS} \\
        \midrule

        \multicolumn{5}{l}{\textit{\textbf{Proprietary Models}}} \\
        GPT-5.2           & 0.9200 & 0.1830 & 0.1930 & 0.3785 \\
        Claude 4.5 Opus   & \textbf{1.0000} & 0.1726 & 0.1179 & 0.3800 \\
        Gemini 3 Pro      & 0.4600 & \textbf{0.4421} & 0.3178 & 0.2969 \\
        Gemini 3 Flash    & 0.8000 & 0.4197 & \textbf{0.4182} & \textbf{0.4592} \\

        \midrule
        \multicolumn{5}{l}{\textit{\textbf{Open-weight Models}}} \\
        DeepSeek-V3.2     & 0.2600 & 0.0237 & 0.0179 & 0.1000 \\
        GLM-4.7           & 0.2200 & 0.0324 & 0.0211 & 0.3200 \\
        Kimi-K2           & 0.5400 & 0.0460 & 0.0214 & 0.1554 \\
        Qwen3-235B        & 0.1200 & 0.0167 & 0.0048 & 0.0600 \\

        \bottomrule
    \end{tabular}
    }
    \caption{\textbf{Stress test on Finance-Ultra-Long setting.}}
    \label{tab:ultra_long_financial}
\end{table}

\begin{figure}[htbp]
  \centering
  \includegraphics[width=\columnwidth]{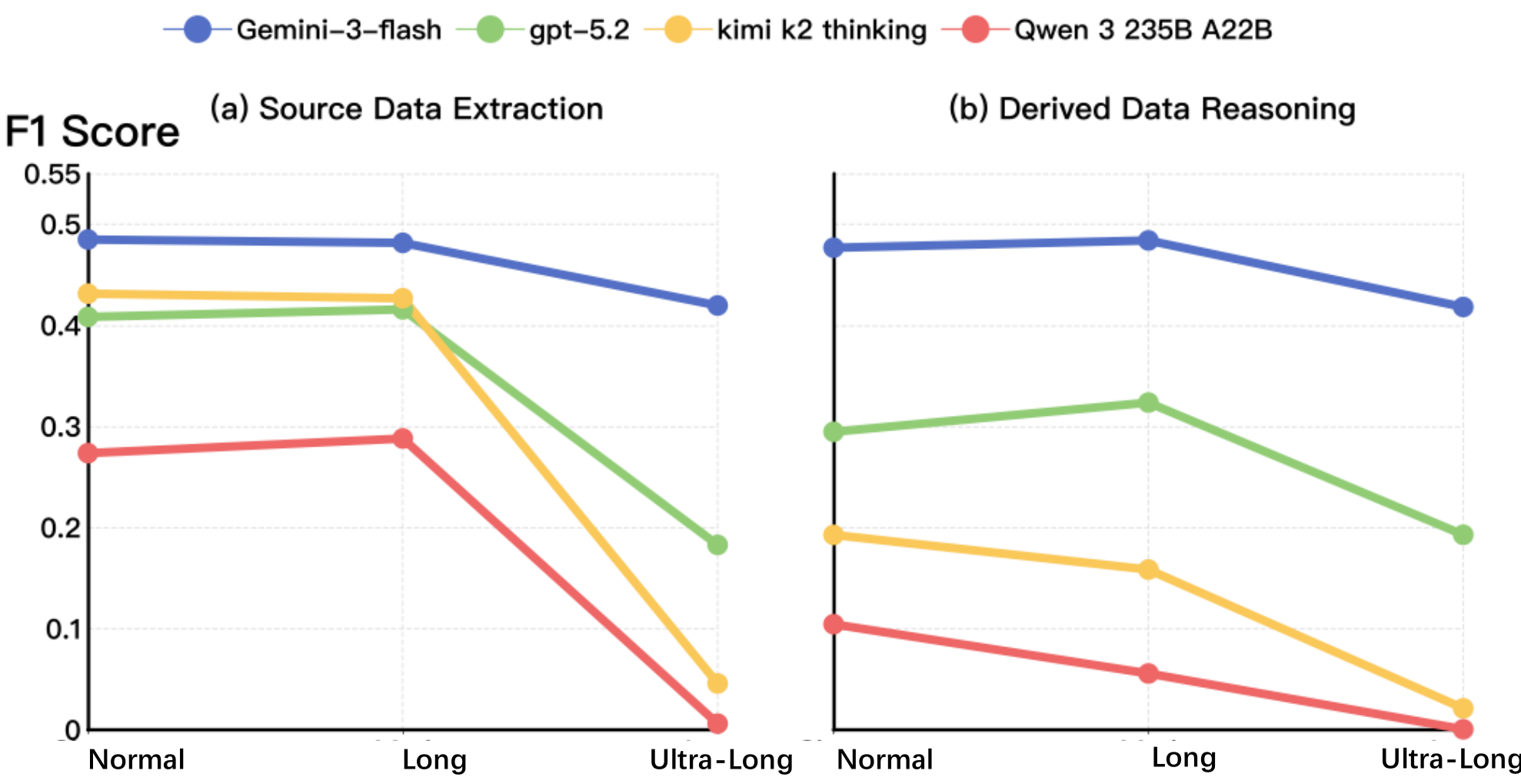}
    \caption{F1 score trends: (a) Source Data Extraction and (b) Derived Data Reasoning as context scales from Normal to Ultra-Long. We visualize 2 proprietary and 2 open-weight models. Detailed results for all eight evaluated models are provided in Table~\ref{tab:ultra_long_financial}.}
  \label{fig:context_impact}
\end{figure}

We evaluate Finance-Ultra-Long as a separate stress test. Beyond Finance-Normal/Long in Table~\ref{tab:text_setting}, this setting provides full 10-K filings as input. Because full-filing inference is costly, we evaluate a fixed 50-instance subset sampled from Finance-Ultra-Long. Appendix~\ref{app:sampling} describes the sampling procedure and representativeness analysis.

Table~\ref{tab:ultra_long_financial} reports the Finance-Ultra-Long results, and Figure~\ref{fig:context_impact} shows how performance changes as Finance contexts scale from Normal to Ultra-Long. Moving from Normal to Long causes only limited fluctuation, suggesting that adding local context around evidence tables has a modest effect. In contrast, Ultra-Long inputs produce a clear drop and larger model divergence. Averaged over all evaluated models, VAS decreases from 0.421 in Normal to 0.269 in Ultra-Long, $F_{1,\mathrm{src}}$ drops from 0.385 to 0.167, and $F_{1,\mathrm{der}}$ drops from 0.249 to 0.139. Full-report financial inputs therefore make both extraction and reasoning substantially harder than table-centered contexts.


The Ultra-Long degradation is consistent with two possible mechanisms. First, full filings may exceed a model’s usable input budget, causing relevant tables or textual cues to be truncated before extraction and reasoning. Second, even when the required evidence is retained, full filings introduce substantial irrelevant context, including tables, notes, and accounting details. This evidence dilution makes it harder to localize sparse task-relevant values and compose them correctly. These results suggest that, in full-document financial settings, longer context windows alone do not guarantee faithful chart generation; models also need robust evidence localization and quantitative reasoning over noisy documents.

\begin{table}[t]
    \centering
    \small
    \setlength{\tabcolsep}{4.5pt}
    \begin{tabular}{lcccc}
        \toprule
        \multirow{2}{*}{\textbf{Model}} & \multicolumn{2}{c}{\textbf{ER}} & \multicolumn{2}{c}{\textbf{VAS}} \\
        \cmidrule(lr){2-3} \cmidrule(lr){4-5}
         & \textbf{Py} & \textbf{HTML} & \textbf{Py} & \textbf{HTML} \\
        \midrule
        \textbf{Proprietary Models} & & & & \\
        GPT-5.2 & 0.765 & 0.983 & 0.521 & 0.573 \\
        Gemini-3-flash & 0.704 & 0.931 & 0.474 & 0.559 \\
        \midrule
        \textbf{Open-weight Models} & & & & \\
        Qwen3-235B & 0.815 & 0.854 & 0.453 & 0.379 \\
        Kimi-k2-thinking & 0.620 & 0.623 & 0.401 & 0.357 \\
        \bottomrule
    \end{tabular}
    \caption{\textbf{Performance comparison between rendering backends.} Data represents the average metrics across Academic-Normal/Long and Finance-Normal/Long.}
    \vspace{-1.\baselineskip}
    \label{tab:modality_comparison}
\end{table}

\subsection{Comparison of Rendering Modalities}

After evaluating the full ERV pipeline, we further isolate the Visualize stage. This allows us to study how rendering backend affects chart faithfulness when the data path is fixed.
Instead of re-running Extract and Reason, we provide each model with the same extracted and derived data $(D_{\mathrm{src}}, D_{\mathrm{der}})$ and evaluate three rendering settings: Python, HTML, and a free-choice setting where the backend is not specified.

Table~\ref{tab:modality_comparison} reports the controlled comparison between Python and HTML. For proprietary models, HTML improves both execution and visual accuracy: average ER increases from 0.735 to 0.957, and average VAS increases from 0.498 to 0.566. For open-weight models, ER remains similar across backends, but average VAS drops from 0.427 with Python to 0.368 with HTML. Thus, backend choice affects not only executability but also visualization faithfulness, and the preferred backend differs across model families.

We further conduct a free-choice setting under the same visualization-only setup, but without specifying the rendering backend. This experiment is performed on the text-domain visualization subset used in Table~\ref{tab:modality_comparison}: for each of the four text settings (Finance-Normal, Finance-Long, Academic-Normal, and Academic-Long), we randomly sample 50 instances and evaluate all eight text-domain models, yielding 1,600 trials in total. Given the same $(D_{\mathrm{src}}, D_{\mathrm{der}})$ input, models choose Python in 1,591 cases (99.4\%), revealing a strong tool-selection bias toward Python. This preference is suboptimal for proprietary models, for which the controlled experiment shows higher VAS with HTML, but it is more consistent with the weaker HTML performance of open-weight models.

\section{Related Work}
\label{sec:related_work}

\paragraph{Chart generation and evaluation.}
Automatic visualization has progressed from recommending or synthesizing charts from structured tables and analytic specifications to LLM-based chart authoring from natural-language instructions, tables, and analysis intents~\citep{hu2018vizmlmachinelearningapproach,dibia2018data2visautomaticgenerationdata,Zhou_2021,Narechania_2021,dibia2023lidatoolautomaticgeneration,wang2023dataformulatoraipoweredconceptdriven,Tian_2025,chen2024visevalbenchmarkdatavisualization}. 
Recent benchmarks such as MatPlotBench, Text2Chart31, C$^2$/ChartUIE, and PlotCraft further evaluate instruction following, executable code generation, and visual rendering quality~\citep{MatPlotBench,pesaran-zadeh-etal-2024-text2chart31,koh2025c2scalableautofeedbackllmbased,zhang2026plotcraftpushinglimitsllms,luo2021nvbenchlargescalesynthesizeddataset,rocco2020efficientneighbourhoodconsensusnetworks}. 
Parallel work on chart understanding, chart-to-code, and chart grounding studies how models read, reproduce, or verify existing charts~\citep{masry2022chartqa,liu2023deplotoneshotvisuallanguage,yang2025chartmimicevaluatinglmmscrossmodal,bansal2025chartabbenchmarkchartgrounding,kafle2018dvqaunderstandingdatavisualizations,methani2020plotqareasoningscientificplots}. 
These studies provide important foundations for visual synthesis. 
However, most settings assume that task-relevant data are already localized, structured, or encoded in an input chart, leaving open whether a generated chart is faithfully grounded in evidence scattered across long documents.

\paragraph{Document-grounded visualization and data analysis.}
Closer to our setting, recent work generates charts or infographics from textual and document contexts, including ChartifyText, Text2Vis, Infogen, Doc2Chart, and DV-World~\citep{zhang2024chartifytextautomatedchartgeneration,jain-etal-2025-doc2chart,ghosh-etal-2025-infogen,rahman-etal-2025-text2vis,meng2026dvworldbenchmarkingdatavisualization}. 
LLM-based data-analysis agents also study data cleaning, code generation, statistical reasoning, tool use, and long-context or multimodal evidence integration~\citep{lai2022ds1000naturalreliablebenchmark,hu2024infiagentdabenchevaluatingagentsdata,huang2024dacodeagentdatascience,hong2024datainterpreterllmagent,jing2025dsbenchfardatascience,li2026longdabenchmarkingllmagents}. 
Related benchmarks on table-text numerical reasoning and document VQA similarly stress evidence retrieval and quantitative reasoning over financial reports, tables, document images, infographics, and slide decks \citep{chen2022finqadatasetnumericalreasoning,zhu2021tatqaquestionansweringbenchmark,tanaka2023slidevqadatasetdocumentvisual}, but they generally return textual or numerical answers rather than executable, evidence-grounded charts.
Yet existing visualization and data-analysis benchmarks usually evaluate either the final visual artifact or the final analysis output.
The intermediate data path from retrieved evidence to chart-ready quantities is rarely made explicit, making it difficult to identify whether failures arise from missing evidence, incorrect derivation, or flawed rendering. 
\textsc{DeepChart} targets this missing intersection: faithful data-science chart generation over long and multimodal contexts.

\section{Conclusion}
We introduced \dname{}, an expert-annotated benchmark for faithful data-science chart generation over task-conditioned contexts derived from real-world scientific papers, financial filings, and ecosystem reports. With an Extract--Reason--Visualize formulation, \dname{} enables stage-by-stage evaluation of extraction, reasoning, and
visualization. Under a zero-shot ERV evaluation protocol, experiments show that current models can produce visually plausible charts while still failing along the
underlying data path, highlighting the need for better evidence localization, quantitative reasoning, and traceable visualization.

\section*{Limitations}

\dname has several limitations. First, although it covers three representative domains---Academic, Finance, and Ecosystem---its scope does not exhaust all data-science chart-generation scenarios. Second, our data-fidelity metrics use normalized value-level matching to accommodate diverse model-generated schemas; this design makes evaluation robust to formatting variation, but the resulting $F1_{\mathrm{src}}$ and $F1_{\mathrm{der}}$ scores should be interpreted as approximate indicators of numerical faithfulness rather than full semantic equivalence. Third, because each instance requires expert verification of source values, derived quantities, executable code, and rendered charts, \dname prioritizes annotation quality and auditability over very large scale. Future work can extend the benchmark to more domains and larger collections while preserving the same verification standard.

\bibliography{ref}

\newpage
\clearpage  

\appendix
\label{sec:appendix}

\section{Benchmark Statistics}
\label{app:benchmark_stats}
\noindent
This appendix summarizes the scale and composition of \textsc{\dname}. 
Table~\ref{tab:statistics} reports the overall benchmark size, including the number of domains, queries, instances, chart types, and query types. 
It also characterizes input scale separately for the text protocol and the multimodal protocol, using token length for text inputs and pages for multimodal inputs.
It also reports average, median, and maximum evidence entries per instance.
Table~\ref{tab:chart-type-distribution} breaks down chart-family coverage by domain, and Table~\ref{tab:query-type-distribution} reports the corresponding distribution of query types.

\FloatBarrier

\begin{table}[!htbp]
\footnotesize
\setlength{\tabcolsep}{3pt}
\renewcommand{\arraystretch}{0.92}
\begin{tabularx}{\columnwidth}{@{}>{\raggedright\arraybackslash}X
                                >{\raggedleft\arraybackslash}p{0.72cm}
                                >{\raggedleft\arraybackslash}p{0.9cm}@{}}
\toprule
\textbf{Domain / Query Type} & \textbf{\#} & \textbf{Share} \\
\midrule
\textbf{Academic} & & \\
- Condition/group comparison & 47 & 26.40\% \\
- Dose/response curve & 27 & 15.17\% \\
- Association/correlation & 20 & 11.24\% \\
- Spatial/geographic comparison & 18 & 10.11\% \\
- Scenario/sensitivity analysis & 16 & 8.99\% \\
- Model performance benchmark & 11 & 6.18\% \\
- Composition/breakdown & 10 & 5.62\% \\
- Temporal trend/projection & 10 & 5.62\% \\
- Flow/network/pathway & 4 & 2.25\% \\
- Genomic/functional mapping & 4 & 2.25\% \\
- Process contribution/LCA decomposition & 4 & 2.25\% \\
- Distribution/variability comparison & 3 & 1.69\% \\
- Regression effect estimate & 3 & 1.69\% \\
- Multivariate profile/clustering & 1 & 0.56\% \\
\midrule
\textbf{Finance} & & \\
- Graham's net-net working capital (NNWC) & 22 & 7.59\% \\
- Earnings quality spread & 22 & 7.59\% \\
- Defensive interval ratio (DIR) & 21 & 7.24\% \\
- EBITDA & 21 & 7.24\% \\
- Piotroski F-Score & 21 & 7.24\% \\
- Return on invested capital (ROIC) & 20 & 6.90\% \\
- Economic value added (EVA) & 20 & 6.90\% \\
- Quality of income ratio & 20 & 6.90\% \\
- Internal growth rate (IGR) & 19 & 6.55\% \\
- Altman Z-Score & 19 & 6.55\% \\
- Reinvestment rate & 19 & 6.55\% \\
- Sloan ratio & 14 & 4.83\% \\
- DuPont return on equity (ROE) & 14 & 4.83\% \\
- Cash conversion cycle (CCC) & 13 & 4.48\% \\
- Cash burn runway & 13 & 4.48\% \\
- Sustainable growth rate (SGR) & 12 & 4.14\% \\
\midrule
\textbf{Ecosystem} & & \\
- Multi-metric relationship & 55 & 21.48\% \\
- Composition and distribution & 28 & 10.94\% \\
- Rank gap and misalignment & 27 & 10.55\% \\
- Growth and momentum & 26 & 10.16\% \\
- Stage and salary & 21 & 8.20\% \\
- Concentration and breadth & 19 & 7.42\% \\
- Capital efficiency & 18 & 7.03\% \\
- Innovation response & 13 & 5.08\% \\
- Funding and deal activity & 12 & 4.69\% \\
- Hierarchy join & 12 & 4.69\% \\
- Exit and liquidity & 8 & 3.12\% \\
- Company quality and maturity & 5 & 1.95\% \\
- Pillar balance & 5 & 1.95\% \\
- Valuation and IPO readiness & 5 & 1.95\% \\
- Composite score & 2 & 0.78\% \\
\bottomrule
\end{tabularx}
\caption{Query-type distributions by domain.}
\label{tab:query-type-distribution}
\end{table}
\begin{table}[!htbp]
\centering
\small
\setlength{\tabcolsep}{3.8pt}
\renewcommand{\arraystretch}{1.03}
\resizebox{\columnwidth}{!}{%
\begin{tabular}{@{}lr@{}}
\toprule
\textbf{Statistic} & \textbf{Number} \\
\midrule
\textbf{Overall} & \\
- Domains & 3 \\
- Queries & 724 \\
- Instances & 1,482 \\
- Chart types & 30 families / 56 specific \\
- Query types & 45 families / 244 specific \\
\midrule
\textbf{Text protocol: tokens / inst.} & \textbf{Avg./Med./Max.} \\
- Academic-Normal. (178) & 5.9K/1.8K/67.2K \\
- Academic-Long (178) & 80.8K/61.7K/166.6K \\
- Finance-Normal. (290) & 35.0K/34.3K/48.0K \\
- Finance-Long (290) & 137.7K/134.3K/197.3K \\
- Finance-Ultra-Long (290) & 706.6K/648.9K/1.2M \\
- overall (1,226) & 220.6K/106.0K/1.2M \\
\midrule
\textbf{Multimodal protocol: pages / inst.} & \textbf{Avg./Med./Max.} \\
- Ecosystem report-level (256) & 218.2/162/402 \\
\midrule
\textbf{Evidence / inst.} & \textbf{Avg./Med./Max.} \\
- Academic & 672.1/176/10.9K \\
- Finance & 82.9/64/225 \\
- Ecosystem & 622.1/400/3.7K \\
- Overall & 317.6/116.5/10.9K \\
\bottomrule
\end{tabular}%
}
\caption{Overall benchmark statistics. Text input scale is measured with the \texttt{cl100k\_base} tokenizer. Multimodal input scale is measured by report pages.}
\label{tab:statistics}
\end{table}

\begin{table}[!htbp]
\centering
\footnotesize
\setlength{\tabcolsep}{3pt}
\renewcommand{\arraystretch}{0.92}
\begin{tabularx}{\columnwidth}{@{}>{\raggedright\arraybackslash}X
                                >{\raggedleft\arraybackslash}p{0.72cm}
                                >{\raggedleft\arraybackslash}p{0.9cm}@{}}
\toprule
\textbf{Domain / Chart family} & \textbf{\#Q} & \textbf{Share} \\
\midrule
\textbf{Academic} & & \\
- Grouped Bar Chart & 43 & 24.2\% \\
- Regression Scatter Plot & 18 & 10.1\% \\
- Dose-Response Plot & 18 & 10.1\% \\
- Heatmap & 12 & 6.7\% \\
- Line Chart & 12 & 6.7\% \\
- Circular/Radial Bar Chart & 11 & 6.2\% \\
- Box Plot & 7 & 3.9\% \\
- Combination Chart & 7 & 3.9\% \\
- Forest Plot & 6 & 3.4\% \\
- Stacked Bar Chart & 5 & 2.8\% \\
- Bar Chart & 5 & 2.8\% \\
- Dot/Point Plot & 4 & 2.2\% \\
- Sunburst/Nested Donut Chart & 4 & 2.2\% \\
- Scatter Plot & 4 & 2.2\% \\
- Violin Plot & 4 & 2.2\% \\
- Sankey/Alluvial Diagram & 4 & 2.2\% \\
- Other types & 14 & 7.9\% \\
\midrule
\textbf{Finance} & & \\
- Bar Chart & 152 & 52.4\% \\
- Line Chart & 138 & 47.6\% \\
\midrule
\textbf{Ecosystem} & & \\
- Horizontal Bar Chart & 78 & 30.5\% \\
- Bubble Chart & 52 & 20.3\% \\
- Dumbbell Chart & 26 & 10.2\% \\
- Ranked Bar Chart & 22 & 8.6\% \\
- Heatmap & 19 & 7.4\% \\
- Slope Chart & 19 & 7.4\% \\
- Scatter Plot & 16 & 6.2\% \\
- Bar Chart & 12 & 4.7\% \\
- Stacked Bar Chart & 7 & 2.7\% \\
- Other types & 5 & 2.0\% \\
\bottomrule
\end{tabularx}
\caption{Chart-family distribution by domain. Types with fewer than four queries are grouped as ``Other types''.}
\label{tab:chart-type-distribution}
\end{table}

\clearpage

\section{Experimental Details}
\label{app:experimental_details}

\subsection{Generation Details}
\label{app:generation_details}
We evaluate model generation with a two-stage pipeline that mirrors the Extract--Reason--Visualize formulation. Given a task context $C$ and chart intent $Q$, the first-stage LLM takes $(C, Q)$ as input and generates an executable Python script, following the extraction-and-derivation prompt shown in Table~\ref{tab:prompt_extraction_derivation}. In this script, the LLM extracts the source values required by the chart from $C$ and encodes them as program variables corresponding to \texttt{src\_data}; it also implements the reasoning operations implied by $Q$ as executable code, which transforms \texttt{src\_data} into chart-ready \texttt{der\_data}. Executing the script writes both fields to a JSON file, yielding the predicted intermediate state $\hat{J}=(\hat{D}_{src}, \hat{D}_{der})$. We then use this predicted intermediate state as the input to a second generation step, where the model is prompted to generate executable visualization code. The generated code is executed to produce the final chart image. This design lets us evaluate not only whether a model can render a chart, but also whether the generated chart is supported by an explicit source-data and reasoning path.

\begin{equation}
\begin{aligned}
\langle C, Q \rangle
&\xrightarrow{\mathcal{M}_{\theta},\,\pi_{\mathrm{ER}}}
\hat{P}_{\mathrm{ER}}
\xrightarrow{\mathrm{Exec}}
\hat{J}=(\hat{D}_{\mathrm{src}}, \hat{D}_{\mathrm{der}}), \\
\hat{J}
&\xrightarrow{\mathcal{M}_{\theta},\,\pi_{\mathrm{vis}}}
\hat{P}_{\mathrm{vis}}
\xrightarrow{\mathrm{Render}}
\hat{G}.
\end{aligned}
\end{equation}

All generation experiments are conducted in a zero-shot setting. We use provider-default decoding parameters unless otherwise specified. For API calls that support reasoning controls, we set \texttt{reasoning\_effort} to \texttt{low}. In text-domain experiments, if the concatenated input exceeds the model-specific context budget, we truncate the context prefix to 80\% of the corresponding context window. In the multimodal Ecosystem domain, each PDF page is rendered at 120 DPI, resized to fit within a $900 \times 1200$ thumbnail, and grouped in original page order into JPEG chunks of up to eight pages each with quality 75. These image chunks are hosted externally and provided to the model as image URLs, with the prompt listing each chunk index and page range. When a report contains more chunks than the model/run-specific input cap, we keep a prefix of the ordered chunks. By default, this cap is set according to the model context window: 32 chunks for models up to 128K tokens, 48 chunks for models up to 256K tokens, 64 chunks for models up to 512K tokens, and 120 chunks for larger-context models; for runs with an explicit cap, we use the specified value and log the selected chunk count.

After each model call, we apply only format-level cleanup to make the output executable, such as removing Markdown code fences or non-code preambles. We do not manually repair generated programs or correct semantic errors in extracted values, derived quantities, or visual encodings. The first-stage Python script is executed with the JSON output path as its command-line argument and a 60-second timeout. A generation is considered to have produced a valid intermediate state only if execution succeeds and the resulting JSON contains non-empty source and derived data fields. The second-stage program is executed with the chart image path as its command-line argument and a 120-second timeout. For HTML outputs used in controlled rendering-backend experiments, we render the page with headless Chromium through Playwright using a $1400 \times 1000$ viewport and save a PNG screenshot after the page has loaded. Executions that time out, crash, or fail to create a non-empty output file are retained as failed generations.

\subsection{Evaluation Details}
\label{app:evaluation_details}

This section supplements the metric definitions in Section~\ref{sec:evaluation} with implementation details. For $F_{1,\mathrm{src}}$ and $F_{1,\mathrm{der}}$, we evaluate the JSON file produced by the first-stage script. Specifically, we recursively extract all numeric leaves from both the predicted and reference JSON objects. Numeric strings are normalized by removing thousands separators and converting them to floating-point values. We then perform one-to-one matching between predicted and reference values. A predicted value is considered correct if its relative error is no larger than $10^{-4}$; for zero-valued references, exact equality is required. Precision, recall, and F1 are computed from these matched values. Missing generated JSON files, invalid JSON files, or outputs without valid source or derived data receive zero data-fidelity score.

This value-level matching intentionally ignores JSON keys and nesting structure. We adopt this design because the benchmark references for \texttt{src\_data} and \texttt{der\_data} are already task-conditioned and contain only the values needed for the target chart, while model-generated JSON schemas vary substantially across models and instances. Requiring exact structural or key-level alignment would therefore penalize formatting choices rather than extraction or reasoning correctness. The one-to-one matching constraint prevents duplicated values from being counted multiple times. When semantic organization errors affect the rendered chart, they are further reflected in the visualization-stage evaluation, especially the data-mapping-topology component of VAS.

For VAS, Gemini 3 Pro generates binary verification rubrics from the reference chart and chart specification. We cache these rubrics once for each benchmark instance and reuse them across all model outputs. The primary VLM judge, \texttt{qwen3-vl-flash-2025-10-15}, answers the rubric items for each generated chart.The rubric items are organized into three aspects: instruction compliance, data-mapping topology, and presentation quality. These items check whether the generated chart satisfies required chart constraints, preserves intended data-to-visual relationships, and remains readable and structurally complete. If a cached rubric is incomplete, we regenerate the missing aspect before evaluation. During evaluation, each generated chart image is paired with each rubric item and sent to the VLM judge, which is required to answer only \texttt{Yes} or \texttt{No}. We set the VLM answering temperature to 0. A satisfied item contributes one point, and an unsatisfied or unclear item contributes zero. The instance-level VAS is the average score over all valid rubric items, and the reported VAS is the macro-average over evaluated instances. If the generated chart image is missing or smaller than the validity threshold, the instance is assigned VAS 0.

Execution Rate is computed from the same execution logs used to collect generated artifacts. A visualization output is marked executable only when the generated program finishes successfully and produces a PNG image larger than the minimum file-size threshold. Python visualization programs are executed with a 120-second timeout. HTML outputs, used only in the rendering-backend analysis, are rendered through headless Chromium and then checked using the same image-validity criterion.

\subsection{Sampling Protocol and Representativeness}
\label{app:sampling}

For the Finance-Ultra-Long setting, we evaluate models on a fixed 50-instance subset to keep inference and evaluation costs manageable. The subset is fixed before model evaluation and is shared by all models and all reported metrics, ensuring a consistent comparison protocol. It covers all 16 financial indicators in the Finance domain and includes both chart families, with 31 bar-chart tasks and 19 line-chart tasks. Its scale and complexity are comparable to the full Finance split: the average input length is 708K tokens, compared with 706.6K tokens for the full 290-instance split; the average number of scalar entries in the reference direct-data records is 86.4 versus 82.9; and the average number of scalar entries in the final chart-ready records is 17.8 versus 17.7. These statistics indicate that the subset preserves the long-context and multi-step financial reasoning characteristics of the full split while making systematic model evaluation feasible.

\subsection{VAS Reliability Validation and Judge Ablation}
\label{app:vas_validation}
To assess the reliability of VAS, we conduct two complementary validation experiments on a fixed 180-chart validation subset. The subset covers the Academic, Finance, and Ecosystem domains, with 60 charts from each domain. Within each domain, we sort all eligible charts by their primary VAS scores from the main experiment and partition them into low, medium, and high strata according to tertiles. We then sample 20 charts from each stratum, ensuring that the validation set covers low-, medium-, and high-quality generations.

\paragraph{Judge-Model Ablation.}

\begin{table}[t]
\centering
\small
\setlength{\tabcolsep}{3.5pt}
\begin{tabular}{lcccccc}
\toprule
Judge Pair & Acc. & F1 & $\kappa$ & $r$ & $\rho$ & MAE \\
\midrule
G5.2 / G4o & 0.776 & 0.702 & 0.415 & 0.650 & 0.392 & 0.162 \\
G5.2 / Q3.5 & 0.898 & 0.878 & 0.755 & 0.875 & 0.778 & 0.078 \\
G5.2 / QVL & 0.773 & 0.711 & 0.427 & 0.722 & 0.513 & 0.129 \\
G4o / Q3.5 & 0.794 & 0.717 & 0.441 & 0.668 & 0.435 & 0.147 \\
G4o / QVL & 0.824 & 0.736 & 0.472 & 0.724 & 0.509 & 0.111 \\
Q3.5 / QVL & 0.778 & 0.708 & 0.418 & 0.685 & 0.453 & 0.128 \\
\bottomrule
\end{tabular}
\caption{Judge-model ablation for VAS. G5.2, G4o, Q3.5, and QVL denote \texttt{gpt-5.2}, \texttt{gpt-4o-mini}, \texttt{qwen3.5-plus}, and \texttt{qwen3-vl-flash-2025-10-15}, respectively. Acc., F1, and $\kappa$ are item-level metrics; $r$, $\rho$, and MAE are chart-level metrics.}
\label{tab:vas_judge_ablation}
\end{table}

We first examine whether VAS depends on a single VLM judge. In this experiment, we keep the binary rubric items fixed and vary only the judge model used to answer the yes/no rubric items. Specifically, we compare \texttt{gpt-5.2}, \texttt{gpt-4o-mini}, \texttt{qwen3.5-plus}, and \texttt{qwen3-vl-flash-2025-10-15}. Each judge receives only the generated chart image and the corresponding rubric items, without access to the generator identity, domain label, primary VAS score, or other judges' outputs.

Table~\ref{tab:vas_judge_ablation} shows that VAS remains reasonably stable across different VLM judges. Across all six judge pairs, item-level agreement is consistently positive, with accuracy ranging from 0.773 to 0.898 and Cohen's $\kappa$ ranging from 0.415 to 0.755. At the chart level, all judge pairs also show positive correlations, with Pearson correlations ranging from 0.650 to 0.875 and Spearman correlations ranging from 0.392 to 0.778. This indicates that different judges generally assign consistent relative VAS scores to the same set of generated charts.

The strongest agreement is observed between \texttt{gpt-5.2} and \texttt{qwen3.5-plus}, which achieve 89.8\% item-level agreement, a Cohen's $\kappa$ of 0.755, a chart-level Pearson correlation of 0.875, and a Spearman correlation of 0.778. Comparisons involving \texttt{qwen3-vl-flash}, the primary judge used in our main experiments, also remain positively correlated with the other judges: its chart-level Pearson correlations with \texttt{gpt-5.2}, \texttt{gpt-4o-mini}, and \texttt{qwen3.5-plus} are 0.722, 0.724, and 0.685, respectively. These results suggest that the VAS trends are not tied to a single judge model, supporting the reliability of our automatic visual evaluation protocol.

\paragraph{Human-Alignment Validation.}
We further validate whether VAS aligns with human visual judgment using the same 180-chart validation subset. The human annotation uses the same binary rubric items as automatic VAS. For each generated chart, annotators are shown only the chart image and its corresponding yes/no rubric items, and are asked to answer each item with \texttt{Yes} or \texttt{No}. Annotators do not see the generator identity, domain label, primary VAS score, automatic judge outputs, or other annotators' answers. Each item is first labeled independently by two human annotators. If the two annotators agree, their shared answer is used as the human reference label; if they disagree, a third annotator adjudicates the item to obtain the final human reference label. This protocol makes the human labels directly comparable to automatic VAS, since both are defined over the same rubric items and the same binary answer space.

We evaluate the agreement between the primary VLM judge and the human reference labels at both the item and chart levels. At the item level, we compare the primary judge's yes/no outputs against the human reference labels and report accuracy, macro-F1, and Cohen's $\kappa$. At the chart level, we aggregate the human binary labels into a human VAS score, computed as the fraction of rubric items answered \texttt{Yes} by the human reference. Automatic VAS is computed analogously from the primary VLM judge's \texttt{Yes} answers. We then report the Pearson and Spearman correlations between automatic VAS and human VAS.

The results show strong agreement between the primary VLM judge and human judgments. Across 2,339 item-level judgments, the primary judge achieves 94.1\% accuracy, 92.5 macro-F1, and a Cohen's $\kappa$ of 0.852 against the human reference labels. At the chart level, automatic VAS is also strongly correlated with human VAS, with a Pearson correlation of 0.939 and a Spearman correlation of 0.878. These results indicate that VAS closely reflects human visual assessment while retaining the scalability of an automatic evaluation protocol.
\clearpage

\onecolumn
\raggedbottom

\section{Prompts}
\label{app:prompts}
Appendix C lists the exact prompts used in \dname. Tables 10-11 define the two-stage generation pipeline, while Tables 12-15 define the automatic VAS rubric generation and judging prompts.

\subsection{Generation Prompts}

\begin{center}
\begin{minipage}{\textwidth}
\begin{PromptBox}{Prompt for Data Extraction and Derivation Script Generation}
\textbf{\textcolor{blue!60!black}{[System Instruction]}}\\
You are an expert Data Engineer. Extract the required data from the given document based on the task requirements, then write a Python script to clean the data and perform the corresponding inference calculations.

\vspace{0.5em}
\textbf{CRITICAL OUTPUT RULE:}
\begin{enumerate}[leftmargin=*, nosep, label=\arabic*.]
    \item Output ONLY raw Python code. No markdown, no explanations.
    \item The code must print the final result as a valid JSON string. JSON only stores clean, processed, and formatted data.
    \item The JSON structure must strictly follow: \texttt{\{"src\_data": "raw values", "der\_data": "derived values"\}}.
    \item The Python script is executed as follows: \texttt{python this.py output.json}. Save this JSON file to a file, and the Python program will accept an argument indicating the path to the output JSON.
\end{enumerate}

\tcbline 

\textbf{\textcolor{blue!60!black}{[User Input]}}\\
Generate a Python script to process data based on the Chart Purpose.

\vspace{0.5em}
\textbf{Chart Type:} \textit{\{\{ chart\_type \}\}} \\
\textbf{Chart Purpose:} \textit{\{\{ chart\_purpose \}\}} \\
\textbf{Chart Layout:} \textit{\{\{ chart\_layout \}\}}

\vspace{0.5em}
\textbf{Logic Requirements:}
\begin{enumerate}[leftmargin=*, nosep, label=\arabic*.]
    \item Extract data relevant to the purpose from the "Documents".
    \item \textbf{Perform any necessary calculations to derive metrics needed for the chart.} 
    \item Only retain the data needed to generate the chart. The \texttt{src\_data} store only the raw data required by the chart, and the \texttt{der\_data} store only the derived data required by the chart. The two data should not overlap.
\end{enumerate}
Save this JSON file to a file, and the Python program will accept an argument indicating the path to the output JSON.

\vspace{0.5em}
\textbf{Documents:}\\
"""\\
\textit{\{\{ data \}\}}\\
"""
\end{PromptBox}
\captionof{table}{The full prompt used for generating Python scripts for data extraction and reasoning. The placeholders in double brackets are replaced with specific test cases.}
\label{tab:prompt_extraction_derivation}
\end{minipage}
\end{center}

\begin{center}
\begin{minipage}{\textwidth}
\begin{PromptBox}{Prompt for Visualization Code Generation}

\textbf{\textcolor{blue!60!black}{[System Instruction]}}\\
You are a specialized data visualization agent. Your task is to analyze the provided \texttt{json\_data}, extract relevant data, and generate strictly executable code.

\vspace{0.5em}
\textbf{CRITICAL OUTPUT RULE:}
\begin{enumerate}[leftmargin=*, nosep, label=\arabic*.]
    \item Output ONLY the raw code. Do not include explanations, summaries, or any text other than the code itself.
    \item If it's a Python file, the program should run as \texttt{python this.py output.png}. Output the chart as a PNG file to a specified location.
    \item If it's HTML, you need to ensure it renders correctly.
\end{enumerate}

\tcbline 

\textbf{\textcolor{blue!60!black}{[User Input]}}\\
Generate raw \textit{\{\{ code\_type \}\}} code to render a \textit{\{\{ chart\_type \}\}} based on the following details:

\vspace{0.5em}
\textbf{Chart Purpose:}\\
\textit{\{\{ chart\_purpose \}\}}

\vspace{0.5em}
\textbf{Chart Layout:}\\
\textit{\{\{ chart\_layout \}\}}

\vspace{0.5em}
\textbf{Input Data (JSON):}\\
\textit{\{\{ json\_data \}\}}

\end{PromptBox}
\captionof{table}{The prompt used for the Visualization Generation module. It takes the structured data (JSON) and layout constraints to produce the final rendering code.}
\label{tab:prompt_vis}
\end{minipage}
\end{center}

\clearpage

\subsection{Evaluation Prompts}

\begin{center}
\begin{minipage}{\textwidth}
\begin{PromptBox}{Prompt for Verification Question Generation}

\textbf{\textcolor{blue!60!black}{[System Instruction]}}\\
You are a data-to-visual mapping verifier. Your job is to write YES/NO verification questions about whether the chart's visual relationships (ordering, trends, relative magnitudes, group comparisons) match what the chart is intended to show.

\vspace{0.5em}
\textbf{CRITICAL OUTPUT RULE:}
\begin{enumerate}[leftmargin=*, nosep, label=\arabic*.]
    \item Output ONLY a single JSON object.
    \item Keys must be "\texttt{q1}", "\texttt{q2}", ... in order.
    \item Values must be ONE English yes/no question each (answerable with "Yes/No").
    \item Do NOT output answers, explanations, or any extra text.
\end{enumerate}

\tcbline 

\textbf{\textcolor{blue!60!black}{[User Input]}}\\
You will be given:
\begin{enumerate}[leftmargin=*, nosep, label=\arabic*)]
    \item A chart image (ground-truth; it satisfies all requirements).
    \item A requirement JSON containing: \texttt{chart\_type}, \texttt{chart\_purpose}, \texttt{chart\_layout}.
\end{enumerate}

\vspace{0.5em}
\textbf{Task:}
\begin{itemize}[leftmargin=*, nosep]
    \item Write 2$\sim$3 YES/NO questions that verify \textbf{DATA MAPPING TOPOLOGY}:
    \begin{itemize}[leftmargin=1em, nosep]
        \item trends (e.g., increasing/decreasing/plateau patterns visible in bars/lines/points)
        \item relative ordering between key conditions (e.g., Condition A > Condition B visually)
        \item correct grouping (individual data points clustered with their corresponding group/bar)
        \item annotation alignment (significance brackets/lines connect the correct groups being compared)
        \item relative magnitude relationships (e.g., largest bar, smallest value positions)
    \end{itemize}
    \item \textbf{IMPORTANT:} Because the image is ground-truth, each question \textbf{MUST} be phrased so that the correct answer for THIS image is "\texttt{Yes}".
    \item Avoid pixel-perfect or exact-value questions; focus on relative/structural correctness.
    \item Questions should be answerable by visual inspection without reading exact axis values.
\end{itemize}

\vspace{0.5em}
\textbf{Requirement JSON:}\\
\textit{\{\{ gen\_parameter \}\}}

\vspace{0.5em}
\textbf{Chart Image:}\\
\textit{\{\{ image \}\}}

\end{PromptBox}
\captionof{table}{The prompt used to generate Ground Truth QA pairs. It instructs the model to create topology-focused verification questions based on a correct reference chart.}
\label{tab:prompt_qgen}
\end{minipage}
\end{center}

\clearpage

\begin{center}
\begin{PromptBox}{Prompt for Compliance Question Generation}

\textbf{\textcolor{blue!60!black}{[System Instruction]}}\\
You are a strict chart-requirements compliance auditor. Your job is to write YES/NO verification questions that check whether a rendered chart obeys explicit constraints.

\vspace{0.5em}
\textbf{CRITICAL OUTPUT RULE:}
\begin{enumerate}[leftmargin=*, nosep, label=\arabic*.]
    \item Output ONLY a single JSON object.
    \item Keys must be "\texttt{q1}", "\texttt{q2}", ... in order.
    \item Values must be ONE English yes/no question each (answerable with "Yes/No").
    \item Do NOT output answers, explanations, or any extra text.
\end{enumerate}

\tcbline 

\textbf{\textcolor{blue!60!black}{[User Input]}}\\
You will be given:
\begin{enumerate}[leftmargin=*, nosep, label=\arabic*)]
    \item A chart image (ground-truth; it satisfies all requirements).
    \item A requirement JSON containing: \texttt{chart\_type}, \texttt{chart\_purpose}, \texttt{chart\_layout}.
\end{enumerate}

\vspace{0.5em}
\textbf{Task:}
\begin{itemize}[leftmargin=*, nosep]
    \item Write 3$\sim$5 YES/NO questions that verify \textbf{INSTRUCTION COMPLIANCE}:
    \begin{itemize}[leftmargin=1em, nosep]
        \item \texttt{chart\_type} constraints (e.g., bar chart + overlaid individual points + error bars)
        \item layout constraints (e.g., single panel, axis arrangement, panel labels, ordering)
        \item purpose-implied required elements (e.g., presence of statistical annotations if significance comparisons are specified)
        \item structural requirements (e.g., grouped vs. stacked, vertical vs. horizontal orientation)
    \end{itemize}
    \item \textbf{IMPORTANT:} Because the image is ground-truth, each question \textbf{MUST} be phrased so that the correct answer for THIS image is "\texttt{Yes}".
    \item Questions must be directly verifiable from the image + the requirement JSON (avoid ambiguous wording).
    \item Focus on binary compliance: either the chart follows the explicit constraint or it does not.
\end{itemize}

\vspace{0.5em}
\textbf{Requirement JSON:}\\
\textit{\{\{ gen\_parameter \}\}}

\vspace{0.5em}
\textbf{Chart Image:}\\
\textit{\{\{ image \}\}}

\end{PromptBox}
\captionof{table}{The prompt used to generate Compliance QA pairs. Unlike the topology prompt, this specifically verifies if the visualization strictly adheres to the user's explicit design constraints.}
\label{tab:prompt_compliance}
\end{center}

\begin{center}
\begin{minipage}{\textwidth}
\begin{PromptBox}{Prompt for Presentation Quality Question Generation}

\textbf{\textcolor{blue!60!black}{[System Instruction]}}\\
You are a visualization presentation-quality inspector. Your job is to write YES/NO verification questions about structural completeness, readability, and clarity of chart elements.

\vspace{0.5em}
\textbf{CRITICAL OUTPUT RULE:}
\begin{enumerate}[leftmargin=*, nosep, label=\arabic*.]
    \item Output ONLY a single JSON object.
    \item Keys must be "\texttt{q1}", "\texttt{q2}", ... in order.
    \item Values must be ONE English yes/no question each (answerable with "Yes/No").
    \item Do NOT output answers, explanations, or any extra text.
\end{enumerate}

\tcbline 

\textbf{\textcolor{blue!60!black}{[User Input]}}\\
You will be given:
\begin{enumerate}[leftmargin=*, nosep, label=\arabic*)]
    \item A chart image (ground-truth; it satisfies all requirements).
    \item A requirement JSON containing: \texttt{chart\_type}, \texttt{chart\_purpose}, \texttt{chart\_layout}.
\end{enumerate}

\vspace{0.5em}
\textbf{Task:}
\begin{itemize}[leftmargin=*, nosep]
    \item Write 3$\sim$5 YES/NO questions that verify \textbf{PRESENTATION QUALITY}:
    \begin{itemize}[leftmargin=1em, nosep]
        \item required elements present (axes, axis labels, units, tick marks, data marks)
        \item readability (text not truncated or overlapping, tick labels are legible)
        \item structural integrity (single coherent panel, consistent alignment, no missing bars/points/lines)
        \item clarity of overlays (error bars, individual points, annotations are visually distinguishable)
        \item proper scaling (axis values increase in correct direction, reasonable tick intervals)
    \end{itemize}
    \item \textbf{IMPORTANT:} Because the image is ground-truth, each question \textbf{MUST} be phrased so that the correct answer for THIS image is "\texttt{Yes}".
    \item Questions must be verifiable from the image; avoid subjective aesthetic preferences (color choice, styling).
    \item Focus on functional quality: can a reader accurately interpret the chart?
\end{itemize}

\vspace{0.5em}
\textbf{Requirement JSON:}\\
\textit{\{\{ gen\_parameter \}\}}

\vspace{0.5em}
\textbf{Chart Image:}\\
\textit{\{\{ image \}\}}

\end{PromptBox}
\captionof{table}{The prompt used to generate Presentation Quality QA pairs. This module focuses on verifying low-level visual correctness, such as element completeness, text legibility, and proper scaling, ensuring the chart is functional and readable.}
\label{tab:prompt_quality}
\end{minipage}
\end{center}

\begin{center}
\begin{minipage}{\textwidth}
\begin{PromptBox}{Prompt for Chart Evaluation Agent}

\textbf{\textcolor{blue!60!black}{[System Instruction]}}\\
You are a precise chart evaluator. Your task is to answer YES/NO questions about chart images based on visual inspection.

\vspace{0.5em}
\textbf{CRITICAL OUTPUT RULE:}
\begin{enumerate}[leftmargin=*, nosep, label=\arabic*.]
    \item Output exactly one token: \texttt{Yes} or \texttt{No} (case-sensitive).
    \item Do NOT output punctuation, prefixes, explanations, reasoning, or newlines.
    \item Answer \texttt{Yes} ONLY if the statement can be clearly and unambiguously verified from the image.
    \item Otherwise (false, missing, or unclear), answer \texttt{No}.
\end{enumerate}

\tcbline 

\textbf{\textcolor{blue!60!black}{[User Input]}}\\
You will be given:
\begin{enumerate}[leftmargin=*, nosep, label=\arabic*)]
    \item A chart image to evaluate.
    \item A specific yes/no question about the chart.
\end{enumerate}

\vspace{0.5em}
\textbf{Task:}
\begin{itemize}[leftmargin=*, nosep]
    \item Examine the chart image carefully.
    \item Answer the given question with "\texttt{Yes}" or "\texttt{No}" based on your visual inspection.
    \item Answer "\texttt{Yes}" only if you can clearly confirm the statement in the question is true.
    \item Answer "\texttt{No}" if the statement is false, or if you cannot clearly verify it from the image.
\end{itemize}

\vspace{0.5em}
\textbf{Question:}\\
\textit{\{\{ question \}\}}

\vspace{0.5em}
\textit{(The chart image is provided as an attached image below.)}

\end{PromptBox}
\captionof{table}{The prompt used for the Evaluation module. The Visual Language Model (VLM) receives the generated chart and a verification question to assess alignment.}
\label{tab:prompt_eval}
\end{minipage}
\end{center}

\clearpage
\section{Representative Instances}
\label{app:examples}

Appendix D shows representative benchmark instances across academic and finance settings. Each example illustrates the context scale, source data, derived data, and final rendered chart required by the ERV pipeline.

\begin{figure*}[!ht]
  \centering
  \includegraphics[width=\textwidth]{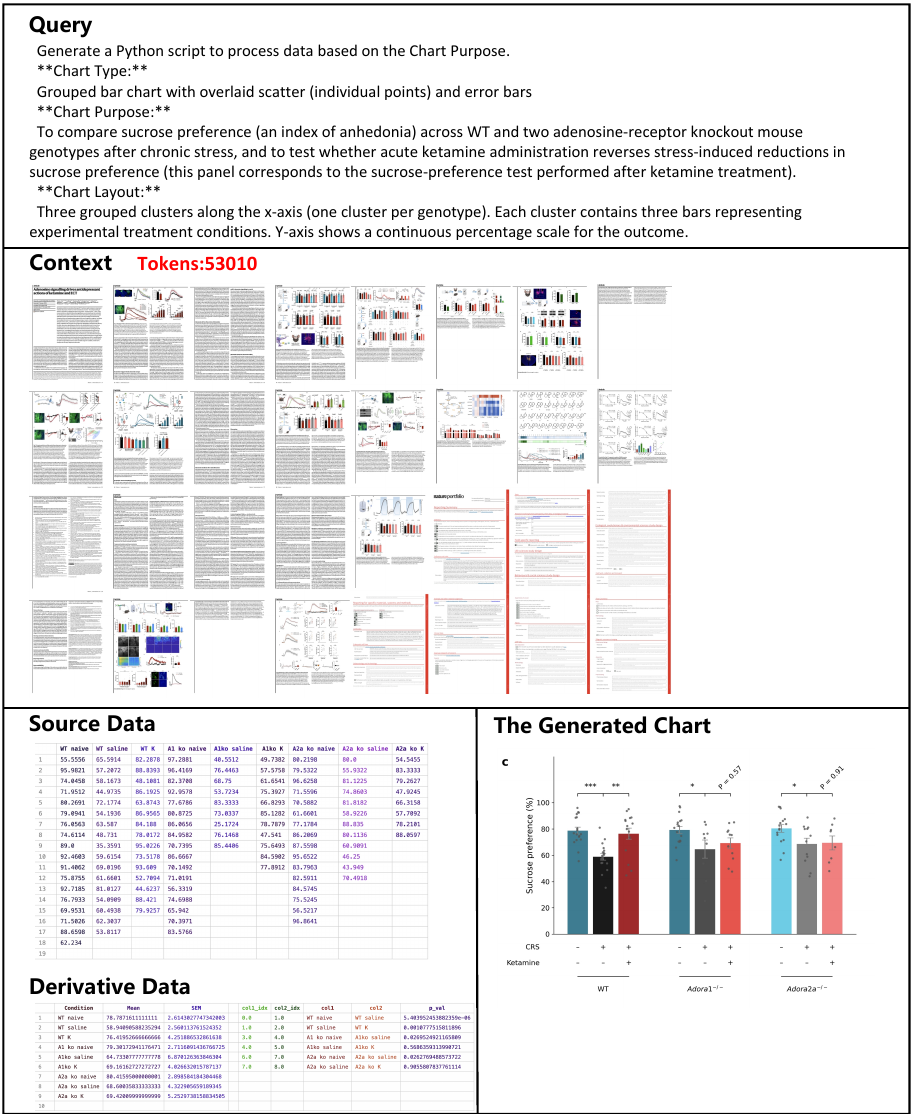}
  \caption{This academic example requires statistical reasoning. The model interprets user intent, extracts raw data points, and infers the underlying statistics, specifically deriving the Mean, Standard Error of the Mean (SEM), and the P-value between conditions.}
\end{figure*}
\newpage
\begin{figure*}[!ht]
  \centering
  \includegraphics[width=\textwidth]{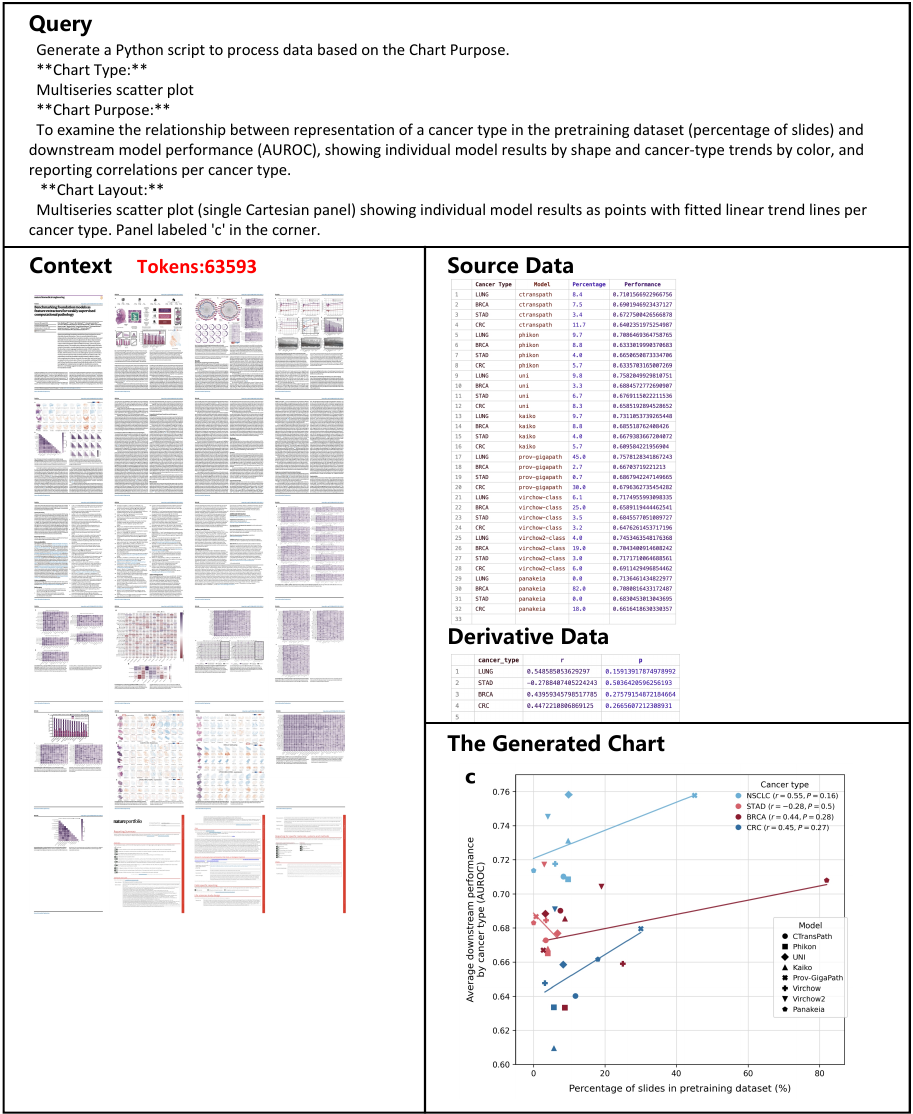}
  \caption{This academic example involves analytical reasoning. The model extracts raw data points to deduce the parameters of each conditional regression line.}
\end{figure*}
\newpage
\begin{figure*}[!ht]
  \centering
  \includegraphics[width=\textwidth]{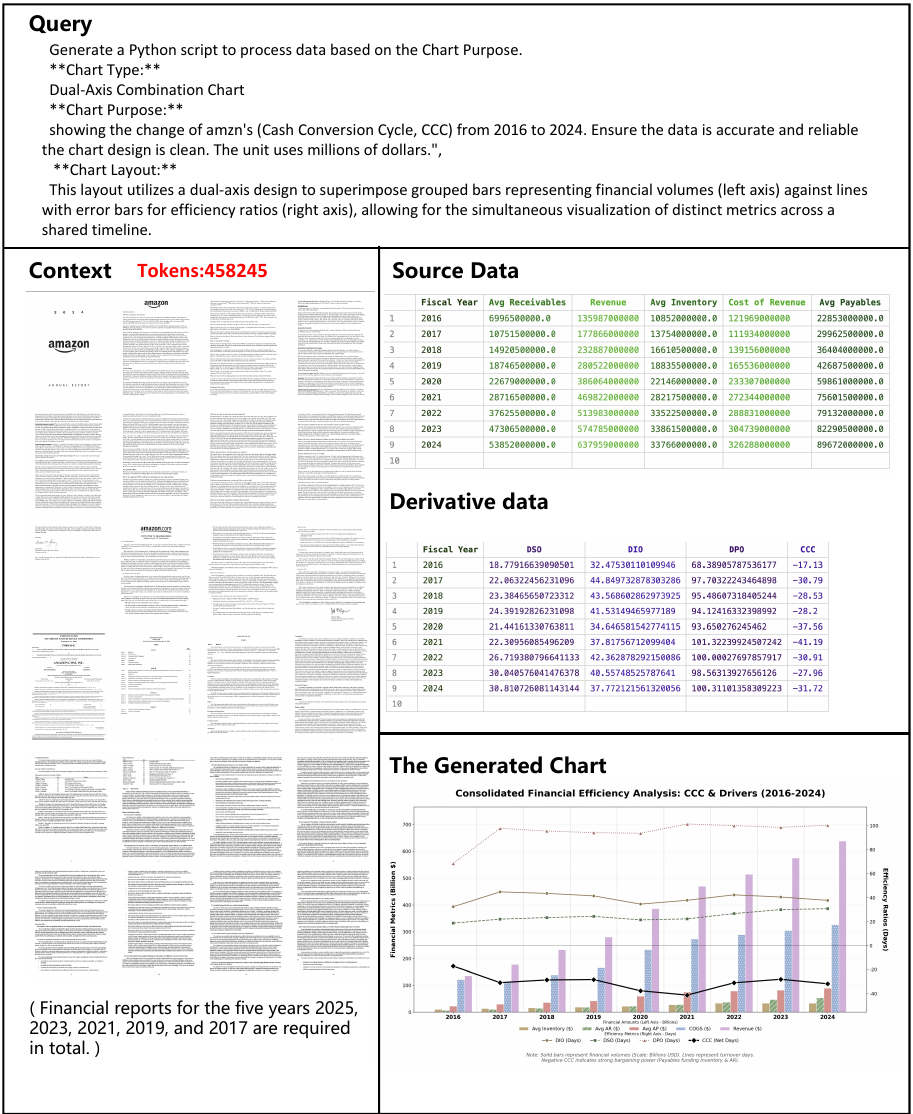}
  \caption{This Finance example demonstrates complex sequential reasoning based on year-end reports (2017 2019 2021 2023 2025). The model extracts base data and executes a multi-hop inference chain: first deriving annual averages, then inferring intermediate metrics (DIO, DSO, DPO), and finally deducing the Cash Conversion Cycle (CCC).}
\end{figure*}
\newpage

\clearpage
\section{Failure Case Studies}
\label{app:bad_cases}
Appendix E provides qualitative failure cases corresponding to the three main failure modes observed in our experiments: incomplete reasoning, temporal coverage mismatch, and incorrect entity selection.

\begin{figure*}[!htbp]
    \centering
    \includegraphics[width=\textwidth]{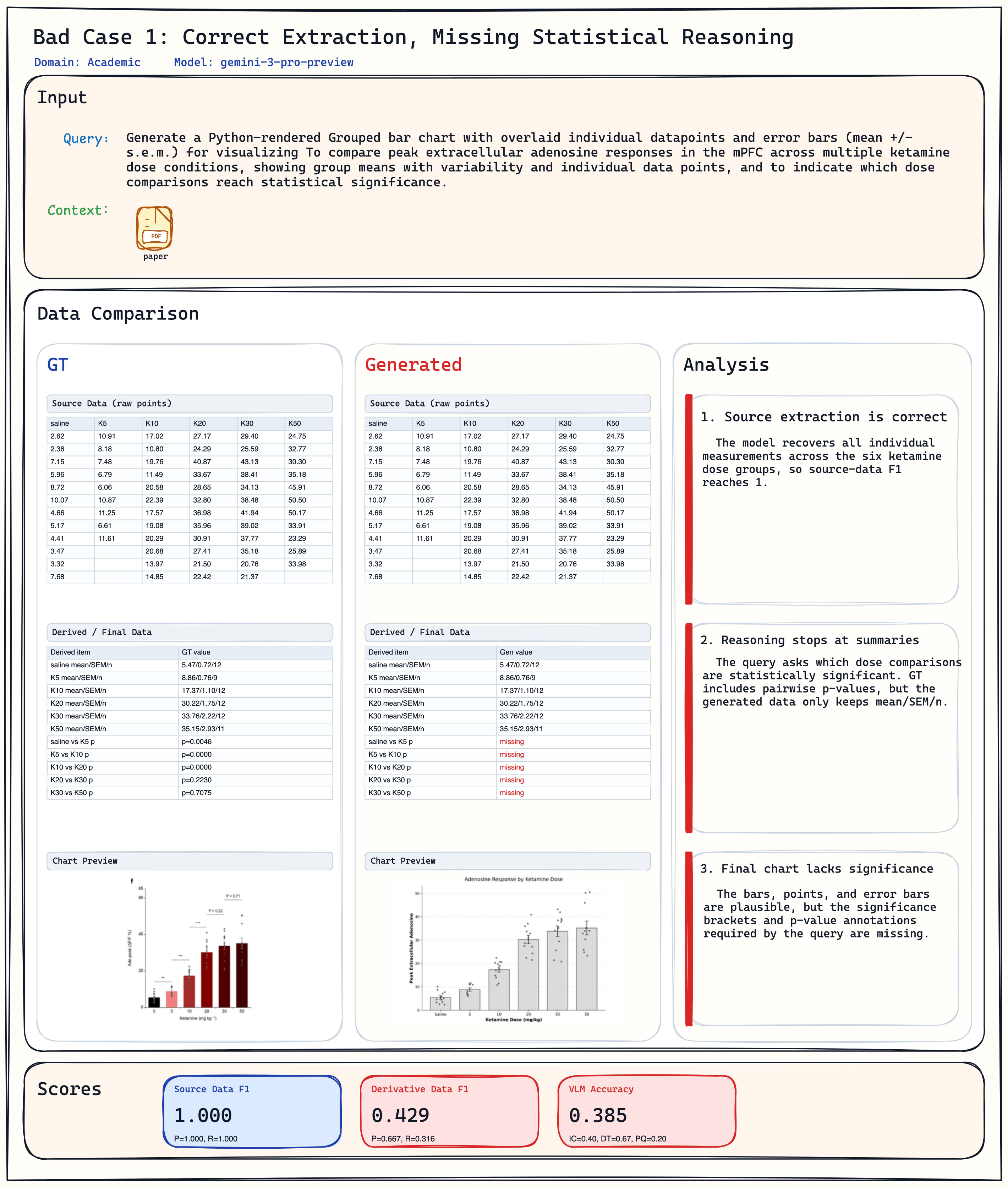}
    \caption{Academic bad case: correct extraction but incomplete statistical reasoning.}
    \label{fig:badcase-finance}
\end{figure*}

\begin{figure*}[t]
    \centering
    \includegraphics[width=\textwidth]{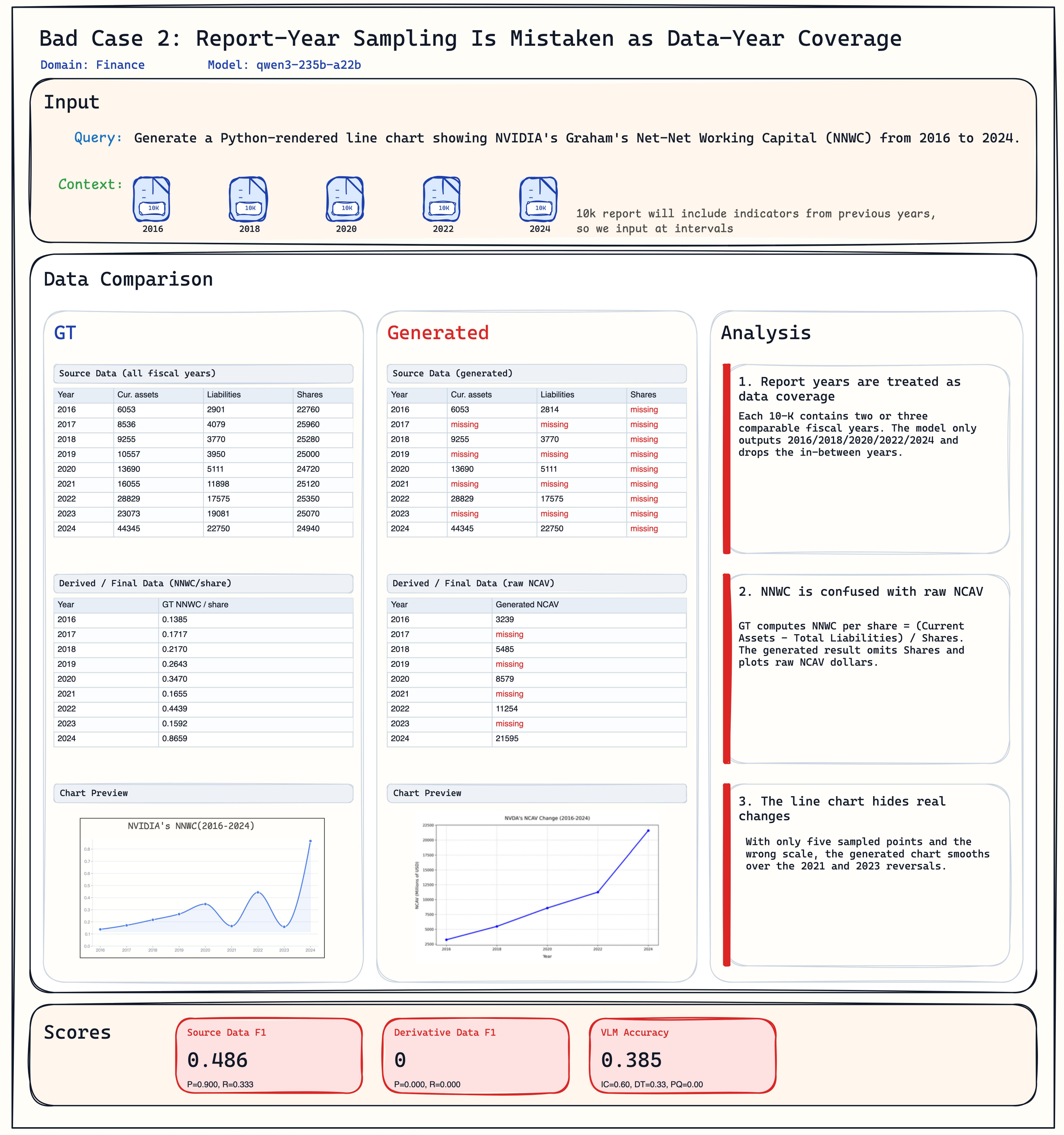}
    \caption{Finance bad case: report-year sampling is mistaken as data-year coverage.}
    \label{fig:badcase-ecosystem}
\end{figure*}

\begin{figure*}[t]
    \centering
    \includegraphics[width=\textwidth]{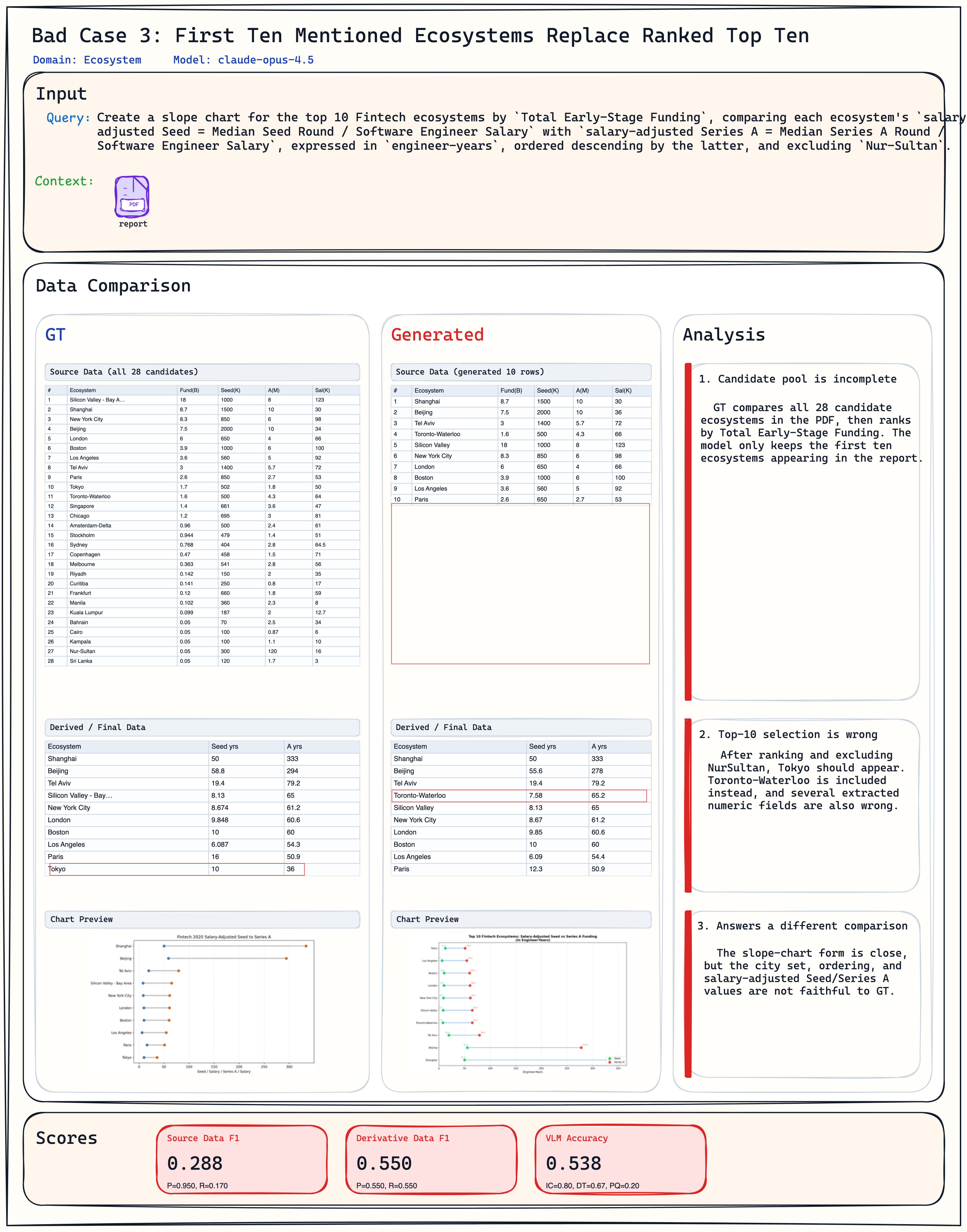}
    \caption{Ecosystem bad case: first-mentioned entities replace the requested ranked top entities.}
    \label{fig:badcase-academic}
\end{figure*}

\end{document}